%% file: acl_latex.tex
\pdfoutput=1

\documentclass[11pt]{article}

\usepackage[final]{acl}

\usepackage{times}
\usepackage{latexsym}
\usepackage{graphicx}
\usepackage{amsmath}
\usepackage{amssymb}
\usepackage{booktabs}
\usepackage{tabularx}
\usepackage{multirow}
\usepackage{pifont}
\usepackage{subcaption}
\usepackage{caption}
\usepackage[normalem]{ulem}
\usepackage{soul}
\usepackage[export]{adjustbox}
\usepackage{colortbl}
\usepackage{makecell}
\usepackage{array}
\usepackage{ragged2e}
\usepackage{xspace}

\usepackage[T1]{fontenc}

\usepackage[utf8]{inputenc}

\usepackage{microtype}

\usepackage{inconsolata}

\input{macros}

\title{Mitigating Open-Vocabulary Caption Hallucinations}

\author{ Assaf Ben-Kish\:\:\: Moran Yanuka\:\:\: Morris Alper\:\:\: Raja Giryes\:\:\: Hadar Averbuch-Elor \\ \\ Tel-Aviv University  \\ \\ \url{https://assafbk.github.io/mocha}}

\begin{document}
\maketitle
\begin{abstract}
   \input{00-abstract}

\end{abstract}

\input{01-intro}

\input{04-openChair}

\input{03-method}

\input{05-exp}
\input{02-rw}

\input{06-conclusion}
\input{07-impact-statement}

\bibliography{egbib}

\appendix

\newpage
\input{supp_00_viz}
\input{supp_01_details}
\input{supp_02_comparisons}
\input{supp_03_extended_rw}

\end{document}

%% file: macros.tex
\newcommand{\hadar}[1]{{\textcolor{magenta}{[Hadar: #1]}}}

\newcommand{\morris}[1]{{\textcolor{cyan}{[Morris: #1]}}}

\newcommand{\ignorethis}[1]{}

\newcommand{\methodname}{\emph{MOCHa}\xspace}
\newcommand{\metricname}{\emph{OpenCHAIR}\xspace}

\makeatletter
\newcommand\footnoteref[1]{\protected@xdef\@thefnmark{\ref{#1}}\@footnotemark}
\makeatother

\newcommand{\XE}{\textsubscript{XE}}
\newcommand{\SC}{\textsubscript{SC}}

\newbox\jsavebox
\newcommand{\jsubfig}[2]{%
	\sbox\jsavebox{#1}%
	\parbox[t]{\wd\jsavebox}{\centering\usebox\jsavebox\\#2}%
	}

\definecolor{bubblegum}{rgb}{0.99, 0.76, 0.8}
\definecolor{corn}{rgb}{0.98, 0.93, 0.36}
\definecolor{cream}{rgb}{1.0, 0.99, 0.82}
\definecolor{bluebell}{rgb}{0.64, 0.64, 0.82}
\definecolor{brilliantlavender}{rgb}{0.96, 0.73, 1.0}
\definecolor{brightube}{rgb}{0.82, 0.62, 0.91}
\definecolor{lavenderindigo}{rgb}{0.58, 0.34, 0.92}
\definecolor{lemonchiffon}{rgb}{1.0, 0.98, 0.8}
\definecolor{lightblue}{rgb}{0.68, 0.85, 0.9}
\definecolor{lightgreen}{rgb}{0.67, 0.88, 0.69}
\definecolor{lightred}{rgb}{0.99, 0.5, 0.5}
\definecolor{awesome}{rgb}{1.0, 0.13, 0.32}

\definecolor{lighterred}{rgb}{0.99, 0.7, 0.7}

\newcommand{\hlc}[2][yellow]{{%
    \colorlet{foo}{#1}%
    \sethlcolor{foo}\hl{#2}}%
}
\newcommand{\hlcr}[1]{\hlc[lighterred]{#1}}

\newcolumntype{R}{>{\raggedright\arraybackslash}X}
\newcolumntype{s}{>{\hsize=.02\hsize}X}

\definecolor{nicegreen}{RGB}{25,170,25}

%% file: 00-abstract.tex
While recent years have seen rapid progress in image-conditioned text generation, image captioning still suffers from the fundamental issue of hallucinations, namely, the generation of spurious details that cannot be inferred from the given image.
Existing methods largely use closed-vocabulary object lists to mitigate or evaluate hallucinations in image captioning, ignoring the long-tailed nature of hallucinations that occur in practice.
To this end, we propose a framework for addressing hallucinations in image captioning in the open-vocabulary setting. %
Our framework includes a new benchmark, \metricname{}, that %
leverages generative foundation models to evaluate open-vocabulary object hallucinations for image captioning, surpassing the popular and similarly-sized CHAIR benchmark in both diversity and accuracy. %
Furthermore, to mitigate open-vocabulary hallucinations without using a closed object list, we propose \methodname{}, an approach harnessing advancements in reinforcement learning. Our multi-objective reward function explicitly targets the trade-off between fidelity and adequacy in generations without requiring any strong supervision. \methodname{} improves a large variety of image captioning models, as captured by our \metricname{} benchmark and other existing metrics. Code and models can be found in:

\vspace{0.5em}
\includegraphics[width=1.45em,height=1.25em]{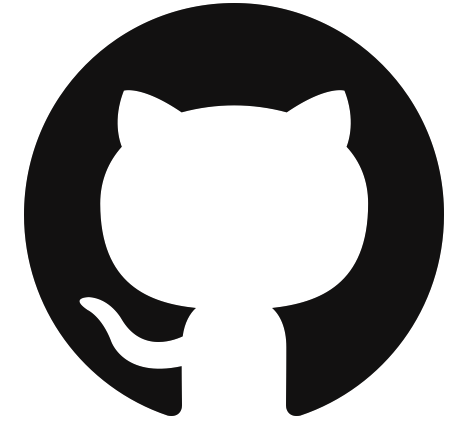}\hspace{.75em}
{\fontsize{8.5pt}{11pt}\selectfont
\parbox{\dimexpr\linewidth-7\fboxsep-7\fboxrule}{\raisebox{0.85em}{\url{https://github.com/assafbk/mocha_code}}}}
\vspace{-.5em}

\input{figures/teaser/teaser}

%% file: figures/teaser/teaser.tex
\begin{figure}
    \includegraphics[trim={4.3cm 3.4cm 1.7cm 2.7cm},clip,width=1\linewidth]{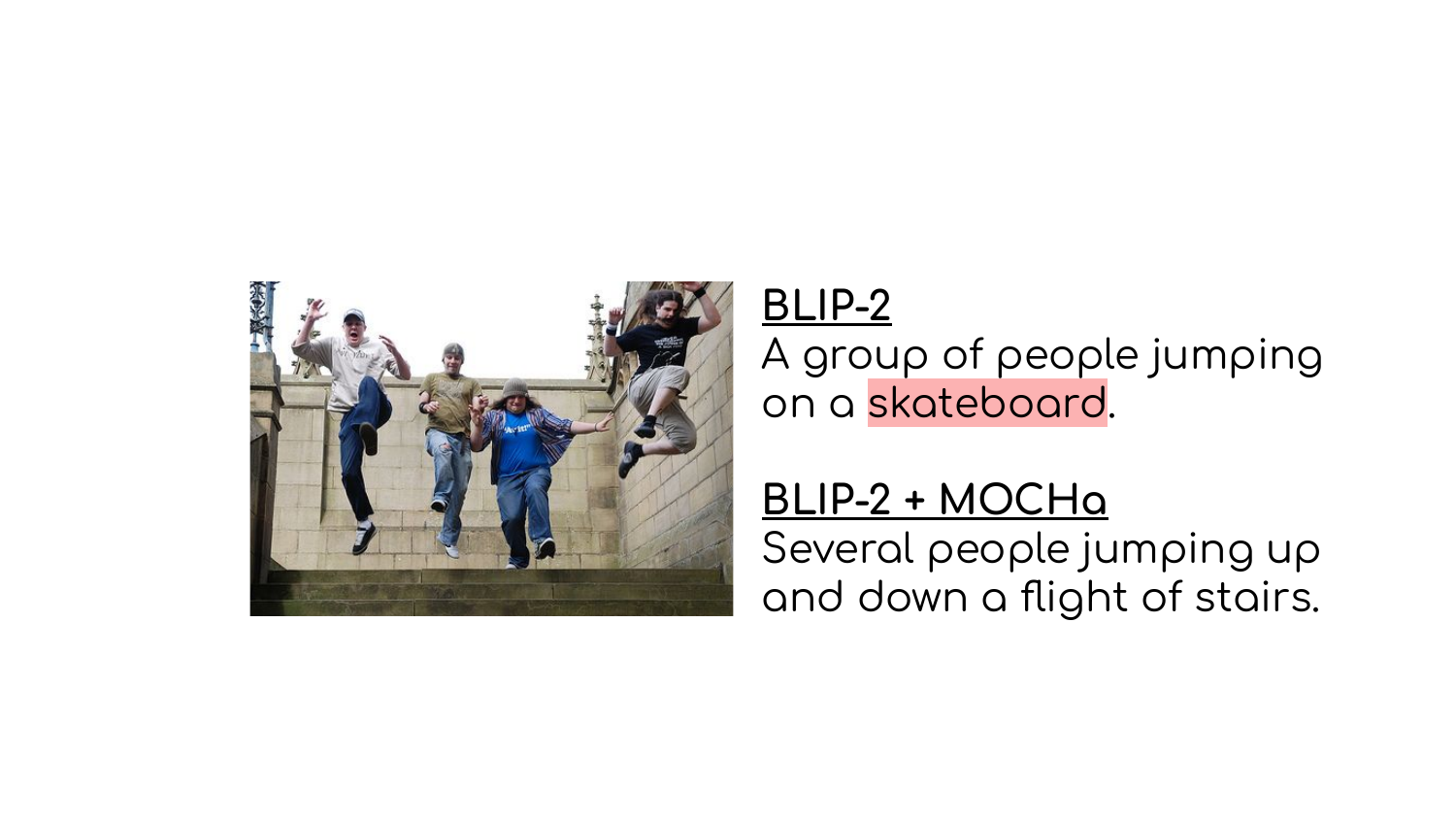}
    
    \vspace{-0.05in}
    \caption{Hallucinated details (shown as \emph{\hlcr{highlighted text}}) are prevalent in the outputs of modern image captioning models, such as the above generation sampled from BLIP2~\cite{li2023blip}. By considering hallucinations in the open-vocabulary setting, we can both quantify and mitigate their effects, illustrated by the improvement provided by our RL-based \methodname{} framework
    (\textbf{+MOCHa}).}
    \vspace{-0.3in}
    \label{fig:teaser}
\end{figure}

%% file: 01-intro.tex
\section{Introduction}
Image captioning, the task of generating text that describes an image, is one of the most fundamental machine learning tasks combining vision and language. Unfortunately, \emph{hallucinations} plague the current state-of-the-art (SOTA), making it less usable for practical tasks that require confidence in the factual correctness of generated captions. Consider, for instance, the image in Figure \ref{fig:teaser}. 
SOTA image captioning models can generate text that is highly semantically related to its associated imagery, but also contains spurious details (``\emph{skateboard}''). Such hallucinated spurious details either damage user confidence or lead to uncritical acceptance of fallacious (and even potentially dangerous) generated content~\cite{chong2022human,mcgowan2023chatgpt,chong2023evolution}.

\input{figures/openchair_scheme/openchair_scheme}

Hallucinations may take a variety of forms in text. %
However, prior work addressing hallucinations in image captioning has largely focused on detecting or mitigating hallucinations by using closed-vocabulary object lists. While this simplifies the problem under consideration, it fails to capture the diversity of hallucinations observed in modern image captioning models. Thus, we propose a framework for both quantifying and mitigating hallucinations in the open-vocabulary setting.

While established benchmarks and metrics for quantifying hallucinations in captioning models exist for closed-vocabulary object sets, they do not exist (to our knowledge) in an open-vocabulary setup. Accordingly, we introduce \metricname{}, a new benchmark for quantifying object hallucinations in an open-vocabulary setting.
We construct our benchmark using text-to-image models and large language models (LLMs) for generating data and performing evaluation. This allows for capturing and accurately quantifying a wide variety of object hallucination types without being limited to a fixed set of categories. Moreover, our open-vocabulary evaluation method considers free-text predictions without referencing a fixed synonym list. Our evaluations show that this outperforms the CHAIR closed-vocabulary metric~\cite{rohrbach2018chair} at capturing performance over diverse hallucinations, providing a complementary measure to CHAIR's evaluation over eighty common object types on natural images.

Equipped with this metric, we turn to hallucination mitigation. A major cause for hallucinations in image captioning are deficiencies in the standard language modeling (LM) objective. 
The \emph{token-level} language modeling objective does not directly optimize the \emph{sequence-level} quality of generated text, and \emph{factual groundedness} is inherently a sequence-level property of text.
Yet, many prior works that directly optimize hallucinations in image captioning limit their scope to a fixed set of possible object tokens, e.g. objects in MS-COCO~\cite{biten2021clockonthebeach,liu2022ciic,petryk2023tlc}, which is incompatible with an open-vocabulary setting.

To mitigate hallucinations without using a closed-vocabulary object list, we introduce \methodname{}, a \textbf{M}ulti-Objective reinforcement learning (RL) based approach for Mitigating \textbf{O}pen-vocabulary \textbf{C}aption \textbf{Ha}llucinations. We observe that RL applied to caption fidelity alone fails to preserve the semantic adequacy (i.e. descriptiveness) of output text, while optimizing for the latter does not enforce factually grounded text. Our key insight is that these two goals can be jointly optimized at the sequence-level by applying RL with a multi-objective reward function. Furthermore, we perform this optimization fully automatically by leveraging SOTA text-based learned metrics, without requiring direct supervision. By considering hallucinations in an open setting, we are able to improve performance across diverse hallucination types, as demonstrated by our \metricname{} benchmark as well as other metrics.
Moreover, we show that our approach can be flexibly applied to a variety of captioning architectures and sizes.

Explicitly stated, our key contributions are:
(i) \metricname{}, a benchmark for open-vocabulary object hallucinations in image captioning.
(ii) \methodname{}, a framework for optimizing a wide array of VLMs to produce high-quality factually-grounded output.
(iii) Experiments showing the advantage of \metricname{} for measuring hallucinations in the open setting, and of \methodname{} for reducing them.

%% file: figures/openchair_scheme/openchair_scheme.tex
\begin{figure*}
    \centering
    \includegraphics[trim={3.2cm 0cm 4.2cm 0cm},clip,width=1\linewidth]{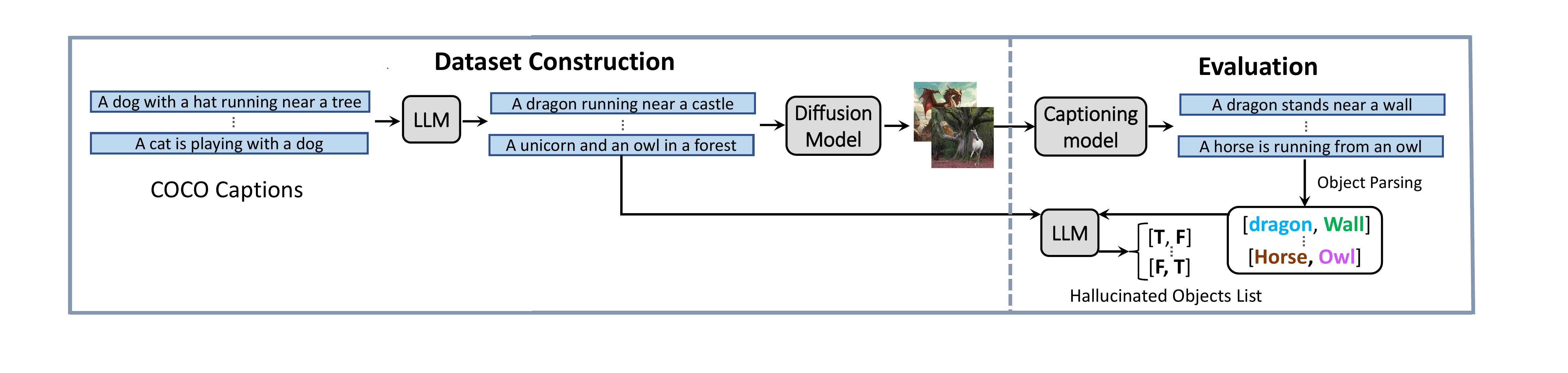}
    \vspace{-0.5in}
    \caption{\textbf{The \metricname{} Benchmark}. We illustrate the construction of the \metricname{} benchmark via an LLM and text-to-image generation model, and its usage for evaluating image captioning models. We first use captions from MS-COCO as seeds to generate diverse synthetic captions. Using syntactic parsing and filtering heuristics, we select for captions containing various open-vocabulary objects. We then generate images corresponding to these captions, producing our benchmark of images linked with object annotations. To evaluate a captioning model, we run it on this benchmark and compare predicted and GT object categories.}
    \label{fig:openchair_scheme}
    \vspace{-0.2in}
\end{figure*}

%% file: 04-openChair.tex
\section{The \metricname{} Benchmark} \label{sec:oc}
To measure object hallucination in the open-vocabulary setting, we propose the \metricname{} (OCH) benchmark, consisting of ${\sim}$5K images illustrating diverse object types in context, accompanied by an evaluation procedure to measure object hallucinations in captioning models. Following existing works~\cite{minderer2022simple,bravo2023open,chatterjee2024opening}, we consider our benchmark to be \emph{open-vocabulary} as it contains diverse and uncommon items reflecting the unlimited distribution found in the real world, as well as having the ability to perform evaluation against arbitrary strings. \metricname{} modifies the previous object hallucination metric CHAIR \cite{rohrbach2018chair}, by relaxing its strong reliance on the object annotations in the MS-COCO dataset, which constitute only 80 common object types.
We control the diversity of object types in our benchmark by leveraging generative models to produce synthetic caption-image pairs, providing a complementary measure to CHAIR's evaluation of a closed set of 80 common objects over natural images. The use of synthetic images for this purpose is further motivated by prior works which show that models training on synthetic image data may generalize to favorable performance on real images~\cite{tian2024stablerep}, as well as the recent growth in usage of synthetic data in general~\cite{sun2024journeydb,betker2023improving}.
We provide an overview of \metricname{} below; further implementation details are provided in the appendix.

In order to create a new benchmark that enables measuring the hallucination rate of arbitrary objects, while still maintaining high quality ground-truth captions, we use the pipeline illustrated in Figure \ref{fig:openchair_scheme}. We first prompt the LLM Llama-2~\cite{touvron2023llama} with few-shot examples of image captions from MS-COCO, having it generate captions with a similar style but containing diverse details (and in particular, objects that are likely not contained in the closed set of MS-COCO object labels). We then parse these synthetic captions with a syntactic parsing model, identify nouns with high concreteness scores~\cite{brysbaert2014concreteness} (as these generally represent concrete objects), and balance the generated captions among object types to cover a wide array of objects. Subsequently, we utilize the text-to-image diffusion model Stable Diffusion XL~~\cite{podell2023sdxl} to generate images from these newly formed captions. This process results in a dataset that consists of synthetic images with corresponding captions including diverse, open-vocabulary objects. While this approach naturally scales to any number of desired image-caption pairs, we generate 5K such pairs (the same order of items found in the widely-used MS-COCO Karpathy test split) and perform manual filtering to assure each pair's alignment and general quality. In total, we removed a small minority (3\%) of generated image-caption pairs.
Figure \ref{fig:openchair_examples} shows examples of image-captions pairs from \metricname{}.

Captioning models may predict free-text objects semantically matching the ground-truth while taking a different surface form (e.g. \emph{chihuahua} vs. \emph{dog}). To capture this in the open-vocabulary setting (rather than using a fixed list of synonyms as done in CHAIR), we evaluate captioning models as follows: After predicting a caption for each image in the \metricname{} dataset, we parse them to identify objects as described above. For each extracted object $o$, we compare it to the ground-truth synthetic caption $c$ by prompting an LLM, asking it whether an image with caption $c$ contains the object $o$ and using its answers to count hallucinations.
Following CHAIR, we calculate the hallucination rate as $n_h / n_{tot}$, where $n_h$ is the number of hallucinated objects (\emph{no} answers) and $n_{tot}$ is the total number of objects considered.
Figure \ref{fig:openchair_demo} illustrates the difference between \metricname{} evaluation and the closed-vocabulary CHAIR metric.

\input{figures/openchair_examples/och_ex}

%% file: figures/openchair_examples/och_ex.tex
\begin{figure}[t]
    \centering

    \jsubfig{\includegraphics[height=2.4cm]{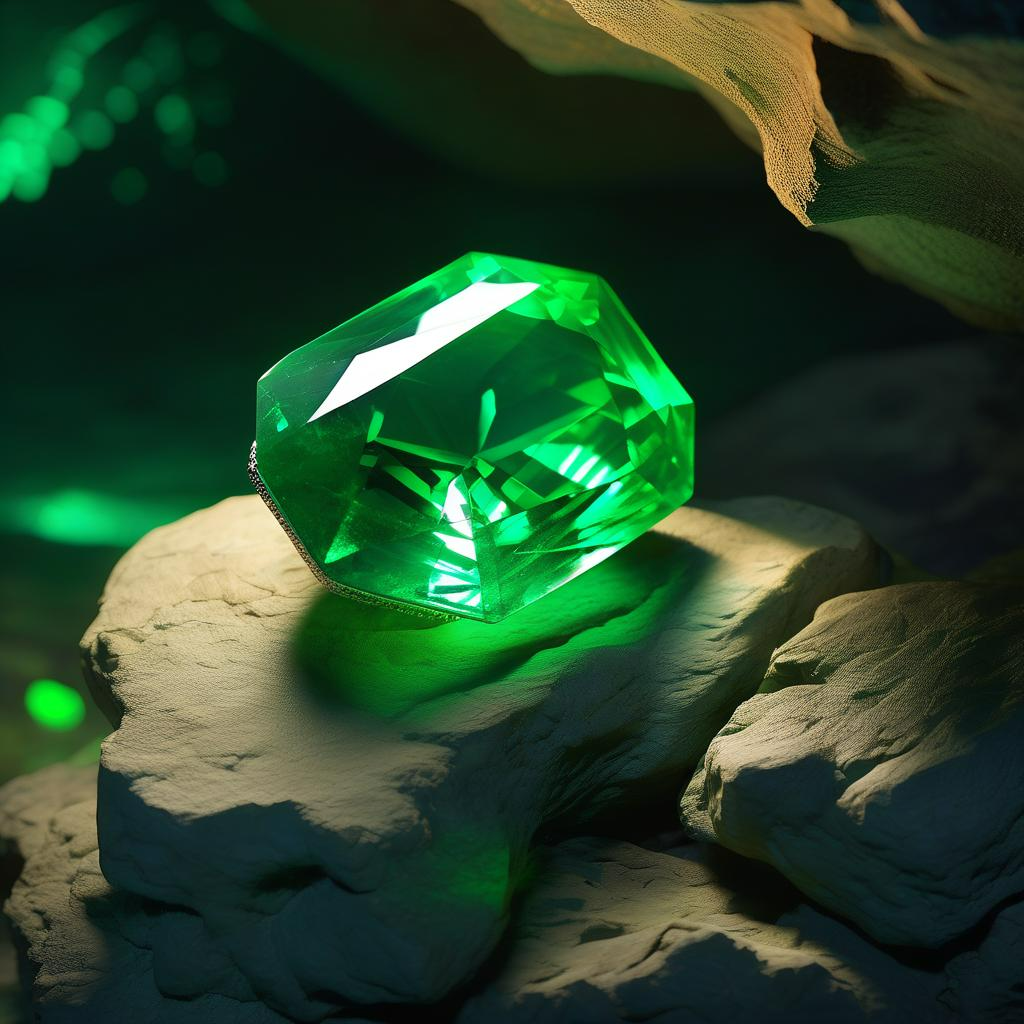}}{{\footnotesize \emph{``A green emerald is perched on a rock in a cave."}}}
    \jsubfig{\includegraphics[height=2.4cm]{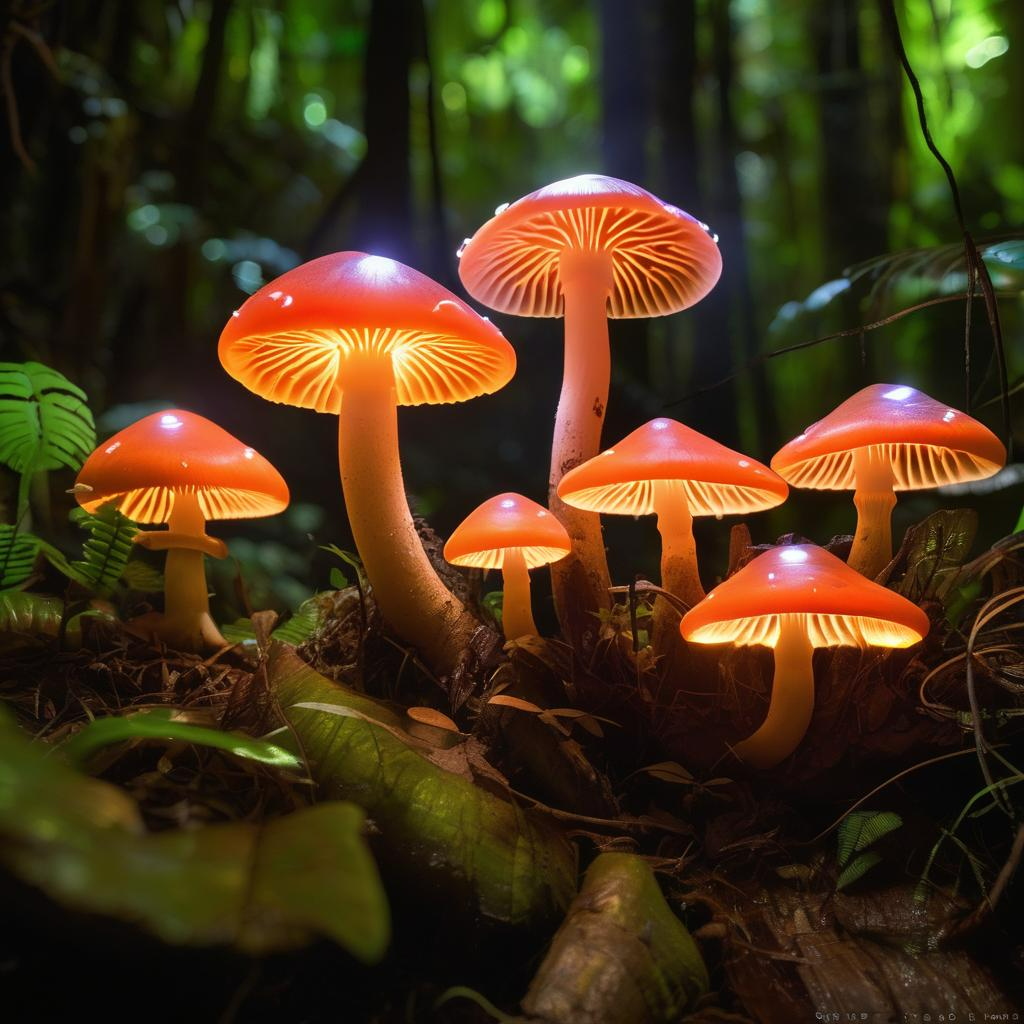}}{{\footnotesize \emph{``A group of mushrooms in the forest."}}}
    \jsubfig{\includegraphics[height=2.4cm]{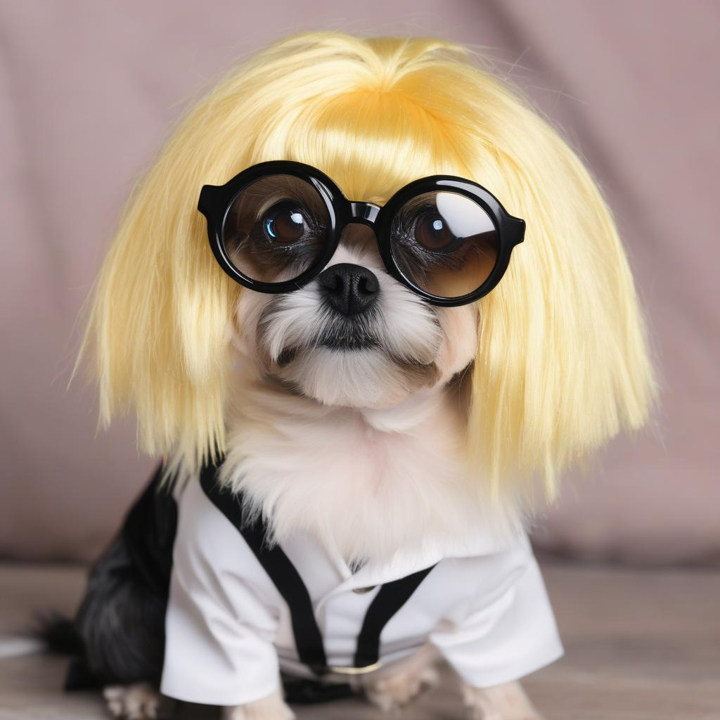}}{{\footnotesize \emph{``A dog dressed as a human with a wig and eyeglasses."}}}    
    \vspace{3pt}

\vspace{-0.1in}
    \caption{\textbf{\metricname{} Examples}. We show examples of images from the \metricname{} benchmark along with their accompanying ground-truth captions, illustrating its diverse coverage of object types.
    Long captions are truncated due to space considerations.}
    \vspace{-0.1in}
    \label{fig:openchair_examples}
\end{figure}

%% file: 03-method.tex
\section{The \methodname{} Framework} \label{sec:method}

To mitigate captioning hallucinations in the open-vocabulary setting, we propose \methodname{}, an RL-based pipeline using SOTA methods for stable reinforcement along with a carefully designed reward function that jointly optimizes for caption fidelity and semantic adequacy. Figure \ref{fig:method_scheme} presents it. 
We turn to describe the learning procedure and objectives used in \methodname{}. We start with preliminaries, then describe the reward function that \methodname{} optimizes (Section \ref{sec:method_reward}), and finally present the RL algorithm used for optimization (Section \ref{sec:method_learning}).

\input{figures/openchair_scheme/openchair_demo}
\input{figures/method_scheme/method_scheme}

\noindent \textbf{Preliminaries.} %
In general, RL views a model as an \emph{agent} that interacts with the external \emph{environment} and receives a \emph{reward}, learning to optimize for this reward via exploring the environment~\cite{Sutton1998}. In the case of image captioning, this model is a VLM operating in an environment of images and reference captions~\cite{rennie2017self}. During training, the agent generates a caption by sampling from its own predicted distribution as shown in Figure \ref{fig:method_scheme} (left), receiving a reward based on an estimate of the caption quality. After collecting a full batch of rewards, a RL optimization step is applied as shown in Figure \ref{fig:method_scheme} (right), and this process repeats iteratively until convergence.

We use the following notation: Let $T$ and $I$ be the sets of possible texts and images, with joint distribution $X$. Given image $i \in I$, an image captioning model $M$ with weights $\theta$ induces a conditional probability distribution $\pi_{\theta}(\cdot | \cdot)$ over generated captions $\hat{c} \in T$ conditioned on images $i \in I$. In the RL context, we refer to $\pi_{\theta}$ as the \emph{policy}. A \emph{reward function} $r : T \times T \times I \to \mathbb{R}$ assigns \emph{reward} (or score) $r(\hat{c}; c, i)$ to generated caption $\hat{c}$ relative to ground-truth caption $c$ and image $i$. %

\subsection{Reward Function} \label{sec:method_reward}
We wish to optimize for the competing objectives of output fidelity (low hallucination rate) and adequacy (including sufficient details to describe the input image), as optimizing for one of these alone causes the other to deteriorate (as shown in our ablations). We also wish to preserve other desired generation properties such as fluency and diversity. To achieve this, we design a reward function combining multiple objectives as follows:

\noindent \textbf{Fidelity Objective.} ($r_f$). To measure output fidelity to the input image, we use the GT reference captions as a proxy, checking for logical consistency via a pretrained Natural Language Inference (NLI) model. This outputs the probability $\overline{p}(\hat{c}, c)$ that the generated text $\hat{c}$ logically contradicts $c$, serving as a strong signal for fidelity, as details which contradict ground-truth information about the image are guaranteed to be hallucinations.
We scale to the range $[-1, 1]$ by using $r_f(\hat{c}; c) := 1-2 \overline{p}(\hat{c}, c)$ as the fidelity reward. We implement this with BART~\cite{lewis2019bartmnli} fine-tuned on the MNLI dataset~\cite{MNLI}. We average values over all reference captions.

\noindent \textbf{Adequacy Objective.} ($r_a$). To measure adequacy (whether the output caption contains sufficient detail), we use BERTScore~\cite{zhang2019bertscore}, a pretrained model measuring text quality relative to ground-truth references. We calculate its F1 value, scaled scale to be approximately in the range $[-1, 1]$ as described in the appendix.

\noindent \textbf{KL Regularization.}
Following prior work~\cite{jaques2017sequence,jaques2019way,ziegler2020finetuning,stiennon2020learning,ouyang2022training}, we add a Kullback–Leibler (KL) divergence penalty to the reward model which constrains the agent to stay close to its initial policy $\pi_0$. This serves to prevent mode collapse (i.e. preserving diversity of outputs) and adversarial policies which over-optimize the reward function. The KL penalty adds a term proportional to $K(\hat{c}; i) := -\log(\pi_{\theta}(\hat{c} | i) / \pi_0(\hat{c} | i))$ to the reward, which limits the agent from excessively distancing itself from the initial policy.

\noindent \textbf{Combined Objective.} Our total reward function takes the form $r(\hat{c}; c, i) := \alpha \cdot r_{f}(\hat{c}; c) + (1-\alpha) \cdot r_{a}(\hat{c}; c) + \beta K(\hat{c}; i)$, where $\alpha \in [0,1]$ and $\beta > 0$ control the trade-off between objectives. %

\subsection{Learning Procedure} \label{sec:method_learning}
To optimize for caption generations that satisfy the desired properties (described above in Section \ref{sec:method_reward}), we adopt the Proximal Policy Optimization (PPO) RL algorithm~\cite{schulman2017ppo}, which has been used by recent works on text generation as discussed in Section \ref{sec:rw}. This is a \emph{policy gradient} algorithm, meaning that it optimizes the parameters $\theta$ in order to (approximately) maximize the expected reward $L(\theta) = E_{i, c \sim X, \hat{c} \sim \pi_\theta(\hat{c} | i)}\left[r(\hat{c}; c, i)\right]$. PPO extends the REINFORCE algorithm~\cite{Sutton1998}, also known as SCST in the context of image captioning~\cite{rennie2017self}, by using a clipped surrogate objective to avoid instabilities.

%% file: figures/openchair_scheme/openchair_demo.tex
\begin{figure}
    \centering
    \includegraphics[trim={1.7cm 5.4cm 0.5cm 3.8cm},clip,width=\linewidth]{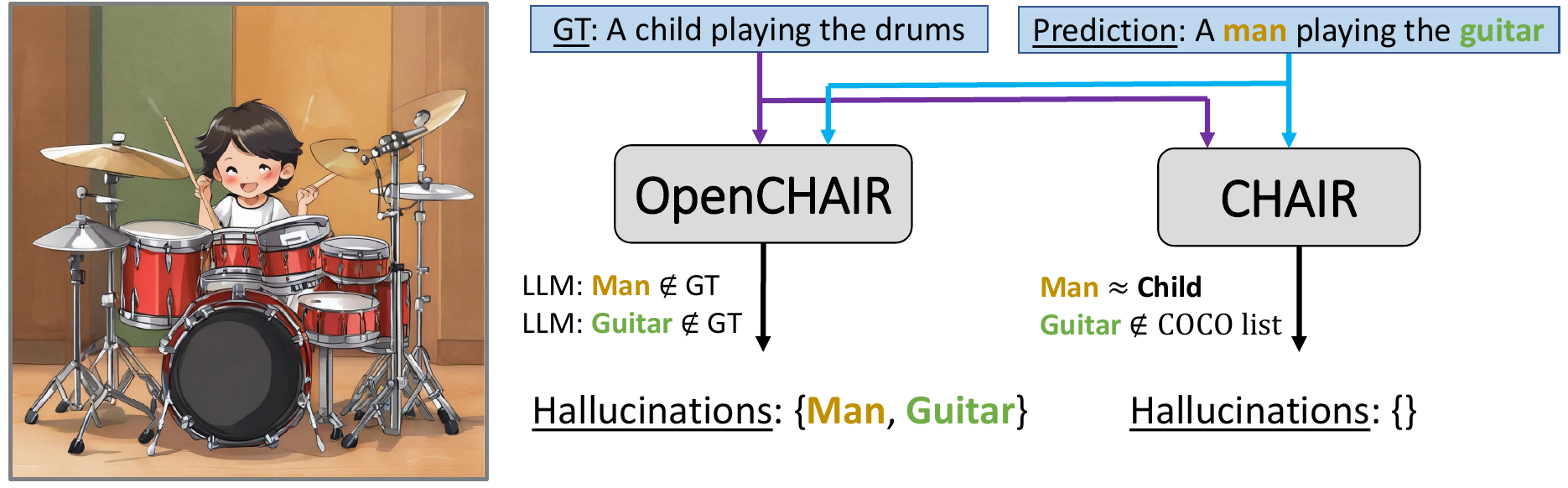}
    \vspace{-20pt}
    \caption{\textbf{\metricname{} vs. CHAIR}.
    In the above the predicted object \emph{guitar} would not be counted by CHAIR since it is not in its fixed vocabulary, while \emph{man} would not be classified as a hallucination since it is defined by CHAIR as a synonym of \emph{child}. In contrast, \metricname{}'s LLM classifies both as hallucinations.}
    \vspace{-0.2in}
    \label{fig:openchair_demo}
\end{figure}

%% file: figures/method_scheme/method_scheme.tex
\begin{figure*}
    \centering
    \includegraphics[trim={24.8cm 3.5cm 3cm 4cm},clip,width=0.95\linewidth]{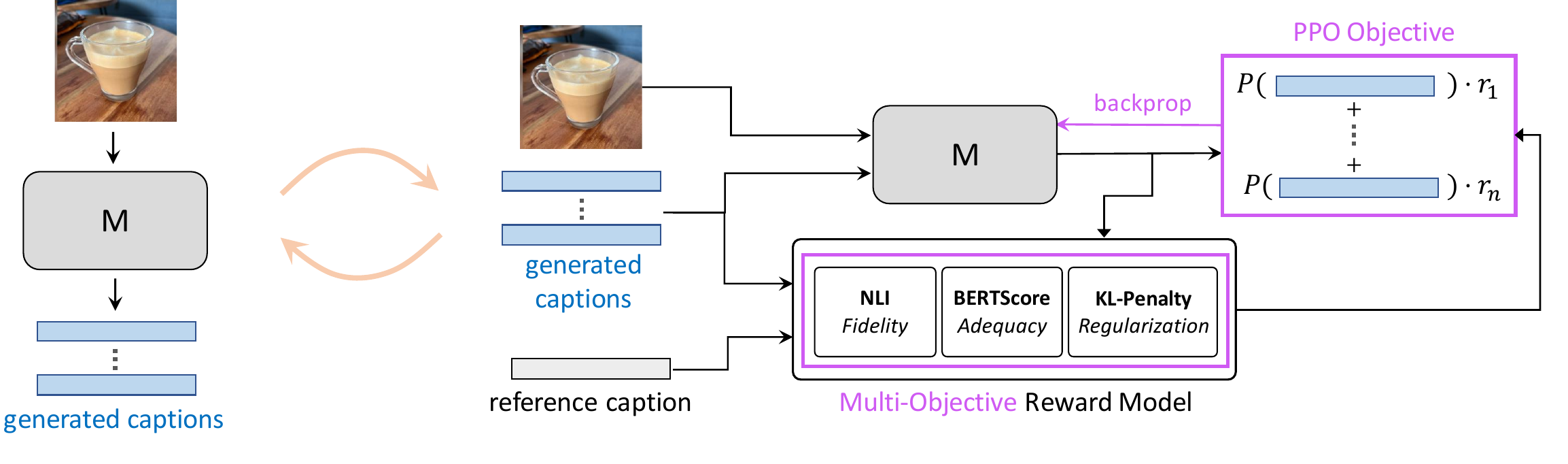}
    \vspace{-0.2in}
    \caption{\textbf{MOCHa scheme}. The algorithm iteratively collects a minibatch of data from an image captioning model $M$ (left side) and then applies an optimization step to the captioning model (right side). The multi-objective reward reinforces $M$ to produce captions closer to the high-scoring captions and further from the low-scoring captions. }
    \vspace{-0.18in}
    \label{fig:method_scheme}
\end{figure*}

%% file: 05-exp.tex
\section{Experiments and Results}
\label{sec:exp}

\noindent \textbf{OpenCHAIR Analysis.}
We analyze the utility of \metricname{} by comparing its distribution of objects to the existing closed-vocabulary CHAIR metric, as well as by performing a human evaluation to compare their correlations to human judgements of hallucinations.

In the first column of Table \ref{tab:och_acc} and in Figure \ref{fig:och_cov} (appendix), we show the difference in the number of unique object types found in CHAIR and \metricname{}, which both contain approximately the same number of images (${\sim}$5K). The open-vocabulary design of \metricname{} enables a significantly larger coverage of object types; in particular, the 2.4K unique object types in \metricname{} reflect an approximately 30-fold increase relative to the 80 object types found in CHAIR. Furthermore, we find that 53\% of object types appear at most three times, and 22\% appear only once, illustrating \metricname{}’s coverage of the long tail of uncommon objects. This is also reflected qualitatively, as the closed-vocabulary benchmark is missing many common object types, including daily objects like \emph{shoe} and \emph{guitar} (see the left image in Figure \ref{fig:chair_issues} for a visual example). In contrast, our benchmark includes diverse object types, such as: \emph{pearl, tiger, sand, tricycle, corkscrew, toy, charcoal, text, pine-cone, grandfather, chocolate, wheelchair, wand, etc.} A large sample of additional objects (those not included in CHAIR) can be found in \url{openchair_objects.txt}. 
Another source of confusion is its synonym list (\emph{e.g.}, see Figure~\ref{fig:openchair_demo}). %

We show that \metricname{} evaluations are grounded in human intuitions via a manual evaluation, comparing its performance to that of CHAIR. For each benchmark (\metricname{} and CHAIR), we generate captions for a random subset of its dataset and manually check object-level decisions (predicted as existing or hallucinated) for over 400 random objects. Results using various captioning models are found in Table \ref{tab:och_acc}. As the presence of hallucinations is highly imbalanced (the large majority of predicted objects are not hallucinated), we report balanced accuracy. We provide further details in appendix \ref{appendix_comparison_och_ch}, including full confusion matrices.

Surprisingly, although operating over a much more diverse scope,  \metricname{} achieves higher accuracy than CHAIR. We identify that this stems from CHAIR's heavy reliance on coarse synonym lists, as seen in Figure \ref{fig:chair_issues} (right). By assessing whether pairs of object names match using a knowledgeable LLM, \metricname{} performs finer-grained hallucination measurements and achieves superior accuracy even in the more general open-vocabulary setting. We note that this reflects a trade-off between true and false positives, as predicted objects may not be found in \metricname{} ground-truth lists despite being present in the accompanying images, due to the limited descriptive capacity of text used to generate images. See more details in the Appendix (Tables~\ref{tab:corr_och_syn} and \ref{tab:corr_ch_coco}).

\input{tables/och_acc}

\input{figures/chair_issues/chair_issues}

\input{figures/main_table/main_table}
\input{figures/qualitative/blip_with_without_mocha_beam_search}
\input{figures/methods/consistency_detail}

As \metricname{} was produced by automatic generation followed by manual filtering, we investigate the effect of the small proportion of erroneous data removed (3\%) on performance. Table \ref{tab:och_filtering_analysis} (appendix) shows that it only marginally impacts the resulting \metricname{} score, validating the high quality of its automatic generation mechanism. Finally, we show that \metricname{} can be calculated more efficiently with smaller LLMs, without compromising evaluation quality. We refer the reader to Appendix \ref{appendx_och_efficiency} for more details. 

\noindent \textbf{\methodname{} Implementation Details.}
We test image captioning with \methodname{} on various SOTA image captioning models of varying architectures and across various sizes. In particular, we test BLIP~\cite{li2022blip}, BLIP-2~\cite{li2023blip} and GIT~\cite{wang2022git}. %
Following standard practice in RL-based image captioning, we use models that have first been fine-tuned on with a standard language modeling loss on the captioning dataset, and then applying PPO reinforcement with our reward function ($\alpha=0.5$). See the appendix for model checkpoints, parameter counts, and further training settings and hyperparameters.

We test our method on the MS-COCO \cite{lin2015mscoco} captioning benchmark, using the data split of Karpathy and Fei-Fei~\cite{karpathy2015deep} (113K items for training, 5K for evaluation). We report standard captioning metrics along with CHAIR~\cite{rohrbach2018chair} and \metricname{} over generated captions (beam search decoding with 5 beams). We also provide NLI ($\overline{p}$) and BERTScore values, directly optimized by \methodname{}, as described in Section \ref{sec:method_reward}. In the appendix, we provide results on additional captioning datasets and metrics to further demonstrate generalization.

\input{tables/mocha_comparisons}
\noindent \textbf{\methodname{} Results.}
Figure \ref{fig:main_table} presents quantitative results of image captioning models on MS-COCO showing the relative improvement of optimizing the baseline SOTA captioning models with \methodname{}. As shown there, \methodname{} improves measures of hallucinations in image captioning while preserving or even enhancing standard measures of caption quality. We note that this is despite the fact that the trade-off between these qualities may degrade one or the other when using a sub-optimal reward weighting (see ablations below). Figure \ref{fig:qual} provides qualitative examples, illustrating that the \methodname{}-optimized model generates captions consistent with the image while preserving a satisfying level of detail, consistent with our numeric results. %

Our quantitative results show that \methodname{} improves performance over base captioning models by most measures, across model architectures and sizes -- not only among metrics that we directly optimize but also among non-optimized metrics, measuring general caption quality (e.g. CIDEr), closed-vocabulary hallucinations (CHAIR) and open-vocabulary hallucinations (\metricname{}). Along with our qualitative observations, this justifies our holistic approach to reducing hallucinations without restriction to a closed object list. Additional results regarding \methodname{} generalization can be found in Appendix \ref{appendix:generalization}.

\noindent \textbf{\methodname{} Comparisons.} %
In Table \ref{tab:comparisons_mocha} we compare \methodname{} to LURE~\cite{Zhou2024Analyzing} and TLC-A~\cite{petryk2023tlc}, current SOTA methods addressing VLM hallucinations, applied to the same pretrained BLIP and BLIP-2 models. LURE fails in the pure image captioning setting as its training procedure encourages long-form, highly detailed outputs. While these are in-distribution for instruction-tuned VLMs, they represent an increase in hallucinations relative to concise captions, as well as an extreme deviation from the reference texts; thus it degrades performance across metrics when applied to captioning models such as BLIP and BLIP-2. Regarding TLC-A, as it targets the objects in the closed-vocabulary object list of CHAIR, it shows an expected advantage in this metric, but does not improve the open-vocabulary hallucination rate (measured by \metricname{}) and even degrades other measures of caption quality, contrasting with the overall improvement shown by our method.
More details and results are provided in Appendix~\ref{sec:lure-details}, \ref{sec:tlc-details} and \ref{sec:add_comp}.

A number of prior works have proposed dedicated methods for reduced-hallucination image captioning, often using data modification or building multi-component pipelines applied to older vision-language backbones. In Table \ref{tab:closed_vocab} (appendix), we provide a comparison between these methods and SOTA foundation VLMs applied as-is, reprodducing results for the dedicated methods UD-L~\cite{biten2021clockonthebeach}, CIIC~\cite{liu2022ciic}, and COSNET~\cite{li2022cosnet}. We find SOTA VLMs outperform these methods across all metrics, motivating our focus on optimization applied on top of modern foundation models.

\noindent \textbf{Ablations.} %
We ablate the components of our reward function, finding that
optimizing for fidelity alone degrades general caption quality, while optimizing for adequacy alone fails to improve hallucinations. This is seen in Figure \ref{fig:fidelity_adequacy} where extreme values of $\alpha$ (0 or 1) correspond to the edges of the curves. Adjusting the parameter $\alpha$ controlling the trade-off between objectives traces a \emph{Pareto frontier} which outperforms the base model, showing that joint optimization of these objectives has a synergistic effect. The effects of each reward function are also illustrated qualitatively in Figure \ref{fig:qual_reward_abl} (appendix); removing $r_f$ from the reward function leads to increased hallucinations, and removing $r_a$ leads to captions that do not contain sufficient details. We provide full numeric results in the appendix, as well as ablating the effect of our chosen RL algorithm and of the KL-Penalty in our reward.

%% file: tables/och_acc.tex
\begin{table}[t]
  \centering
  \setlength{\tabcolsep}{3pt}
    \begin{tabular}{l|c|cccc}
        \toprule
              & \multirow{2}{*}{\shortstack{\# Obj\\ Types}} & \multicolumn{4}{c}{Balanced Accuracy}  \\
              &      &     BLIP2  &   BLIP-L  & GIT-B &  OFA-L   \\
        
        \midrule
        
        CH    & 80*   &  0.844      & 0.774      & 0.899   & 0.810     \\

        OCH   & \textbf{2400} &  \textbf{0.945}      & \textbf{0.944}     & \textbf{0.943}   & \textbf{0.930}     \\

        \bottomrule
        
    \end{tabular}

\vspace{-5pt}
  \caption{\textbf{Human Evaluation of \metricname{} and CHAIR}.
  We perform a manual evaluation of \metricname{} and CHAIR object-level predictions, as described in Section \ref{sec:exp}. As seen above, \metricname{} covers a much larger variety of unique object types while also outperforming CHAIR in per-object predictive accuracy (of whether the given object is present or hallucinated).
  *CHAIR includes also a synonym list.
  }
  \vspace{-0.1in}
  \label{tab:och_acc}
\end{table}

%% file: figures/chair_issues/chair_issues.tex
\begin{figure}
    \centering
    \includegraphics[trim={2.7cm 5.2cm 8.7cm 0.4cm},clip,width=0.95\linewidth]{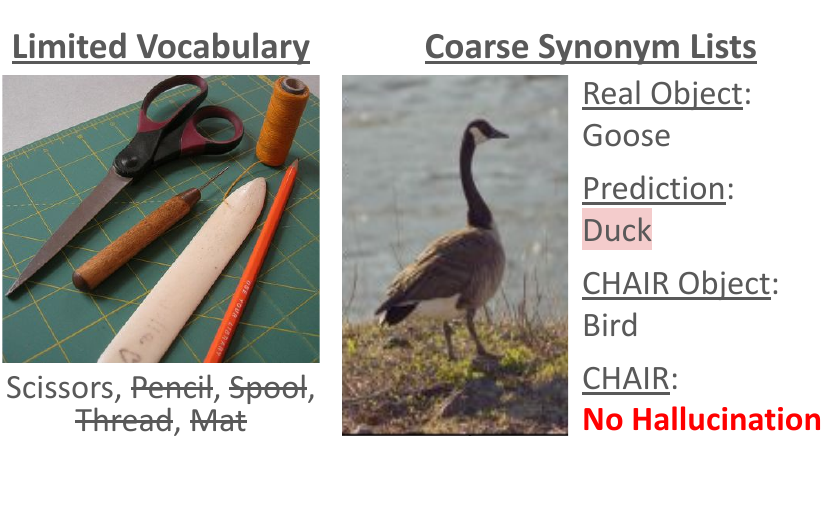}
    \vspace{-0.25in}
    \caption{\textbf{CHAIR Limitations.} The left image exhibits CHAIR's limited vocabulary. Out of all objects predicted by BLIP2, \textit{Scissors} is the only object CHAIR considers during the evaluation. The right image illustrates a limitation stemming from CHAIR's use of a fixed list of synonyms to coarsely aggregate different, semantically similar objects. Hallucinations that occur within the same synonym group are considered as a correct detection; in this example both \textit{Goose} and \textit{Duck} are defined as synonyms of \textit{Bird} even though the image does not display a duck (but rather a goose).
 }
    \label{fig:chair_issues}
    \vspace{-0.2in}
\end{figure}

%% file: figures/main_table/main_table.tex
\begin{figure*}[t]
    \vspace{-0.1in}
    \centering
    \includegraphics[trim={0cm 2.6cm 0cm 4.4cm},clip,width=1\linewidth]{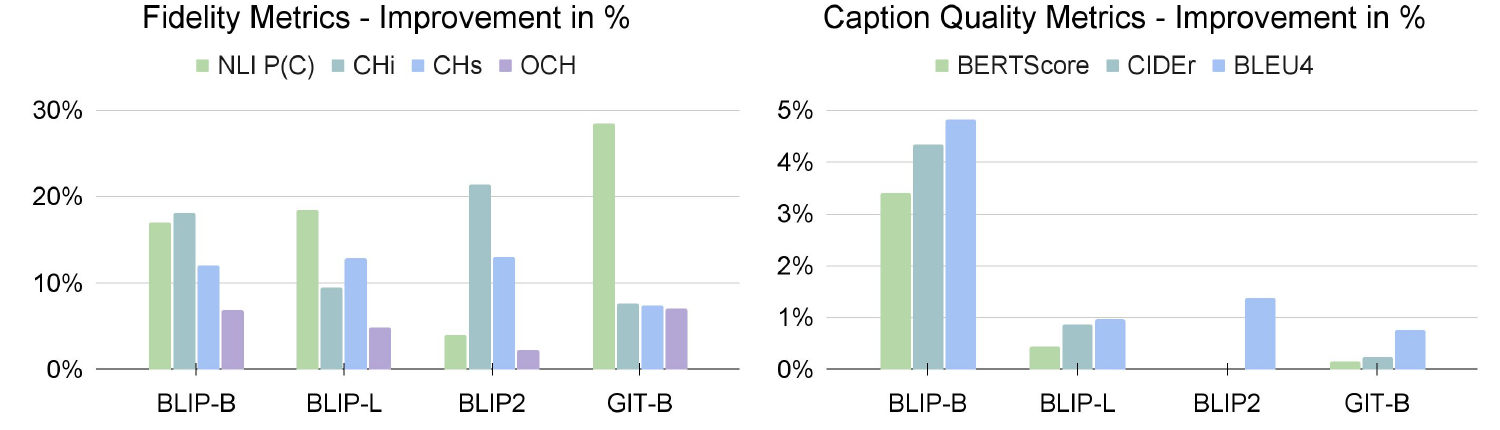}
    \vspace{-0.27in}
    \caption{\textbf{Reducing Hallucinations While Maintaining Caption Quality}. We show the relative improvement of state-of-the-art VLM models when optimized using \methodname{} optimization on the COCO Caption Karpathy test set. CH and OCH refer to Chair and \metricname{} respectively. All results are generated by using their officially provided checkpoints and hyperparameters.
    Full numeric results are provided in the appendix.%
    } 
    \vspace{-0.15in}
    \label{fig:main_table}
\end{figure*}

%% file: figures/qualitative/blip_with_without_mocha_beam_search.tex
\begin{figure}[t]
\centering
\setlength{\tabcolsep}{1.1mm}
\begin{tabular}{>{\centering\arraybackslash}p{0.08\linewidth} |  p{0.255\linewidth} | p{0.255\linewidth} | p{0.255\linewidth} p{0.001\linewidth}}

\raisebox{0.025\textwidth}{\parbox{1cm}{\centering \textbf{ }}} & 
\includegraphics[trim={2.5cm 0cm 2.5cm 1.5cm},clip,width=2cm,height=1.86cm]{figures/qualitative/beam_search/1394_suit_military.png} & 
\includegraphics[trim={2cm 0cm 3.2cm 0cm},clip,width=2cm,height=1.86cm]{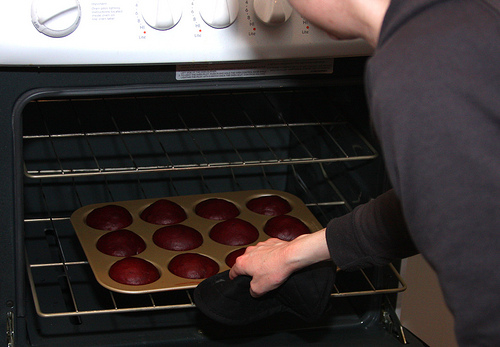} & 
 \includegraphics[trim={3cm 1cm 1cm 0cm},clip,width=2cm,height=1.86cm]{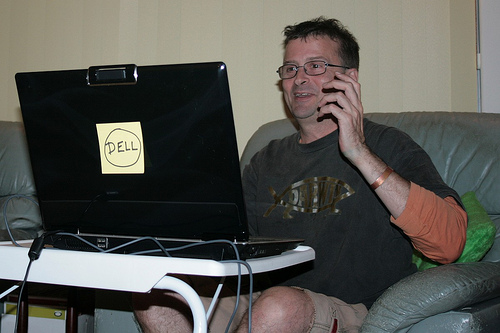}  \\ \hline

\raisebox{-0.03\textwidth}{\small{\textbf{B}}} & \footnotesize{\emph{A man in a suit and tie standing by \hlcr{another man in a suit and tie}}} & \footnotesize{\emph{A person taking a tray of \hlcr{apples} out of an oven}} & \footnotesize{\emph{A man sitting on a couch \hlcr{talking on a cell phone}}} & \\ \hline

\raisebox{-0.03\textwidth}{\small{\textbf{B+M}}} & \footnotesize{\emph{A man in a military uniform talking to a man in a suit and tie}} & \footnotesize{\emph{A person taking a pan of food out of an oven}} & \footnotesize{\emph{A man sitting on a couch using a laptop computer}} \\

\end{tabular}

\vspace{12pt}

\centering
\setlength{\tabcolsep}{1.1mm}
\begin{tabular}{>{\centering\arraybackslash}p{0.08\linewidth} |  p{0.255\linewidth} | p{0.255\linewidth} | p{0.255\linewidth} p{0.001\linewidth}}

\end{tabular}

\vspace{-18pt}
\caption{\textbf{Qualitative results} of \methodname{} applied to an image captioning model (BLIP-Large), along with baseline results without optimization (noted as B+M, B, respectively). We show captions (over COCO) produced from each model using beam search decoding with five beams.
Hallucinated details are \hlcr{highlighted}. The results illustrate that \methodname{} encourages captions with high fidelity to the input image (avoiding hallucinations), while preserving a satisfying level of detail.}
\label{fig:qual}
\vspace{-0.2in}
\end{figure}

%% file: figures/methods/consistency_detail.tex
\begin{figure}[t]
    \centering
    \jsubfig{\includegraphics[height=3.8cm]{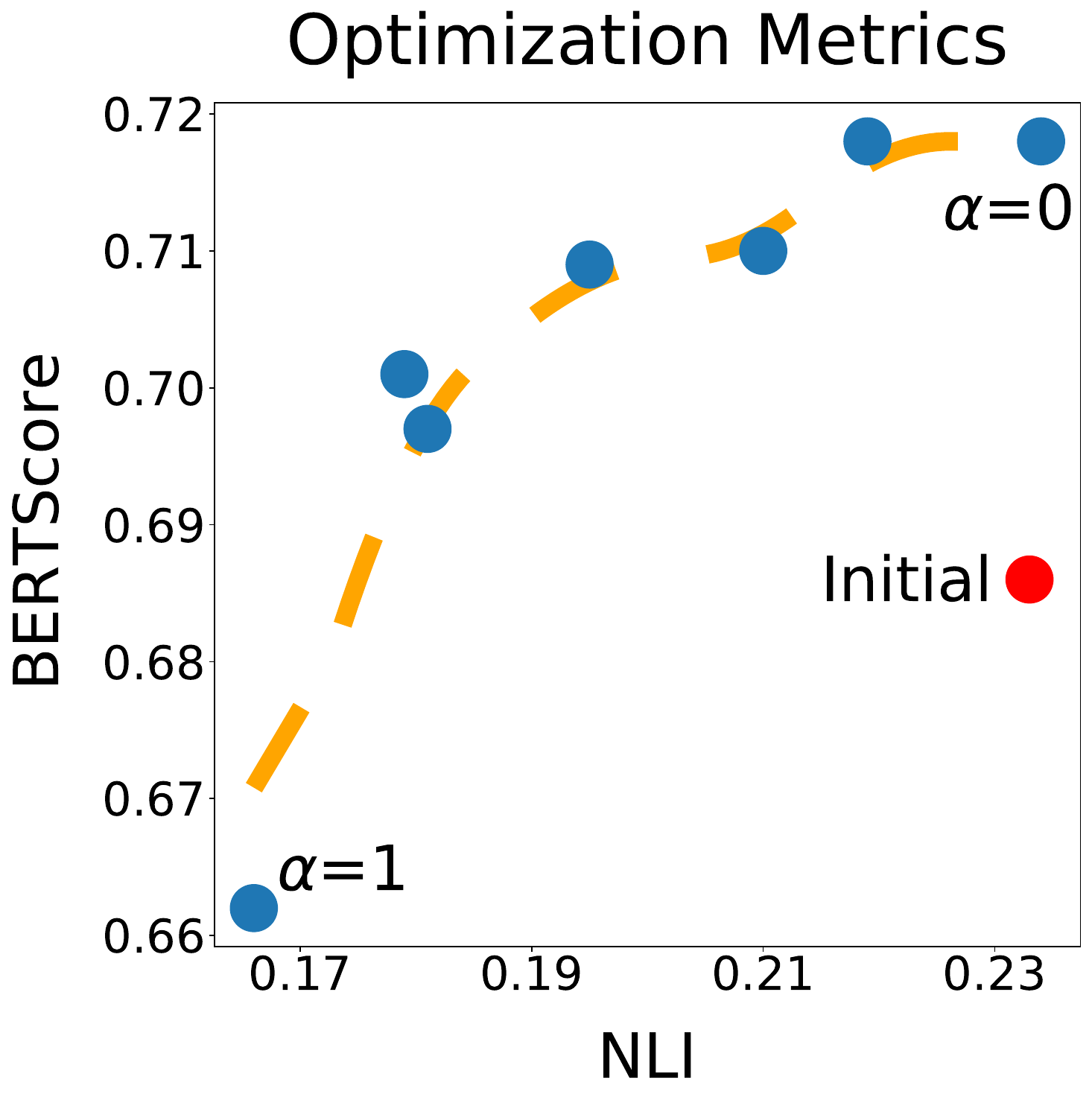}}{}
    \hfill
    \vspace{-10pt}
    \jsubfig{\includegraphics[height=3.8cm]{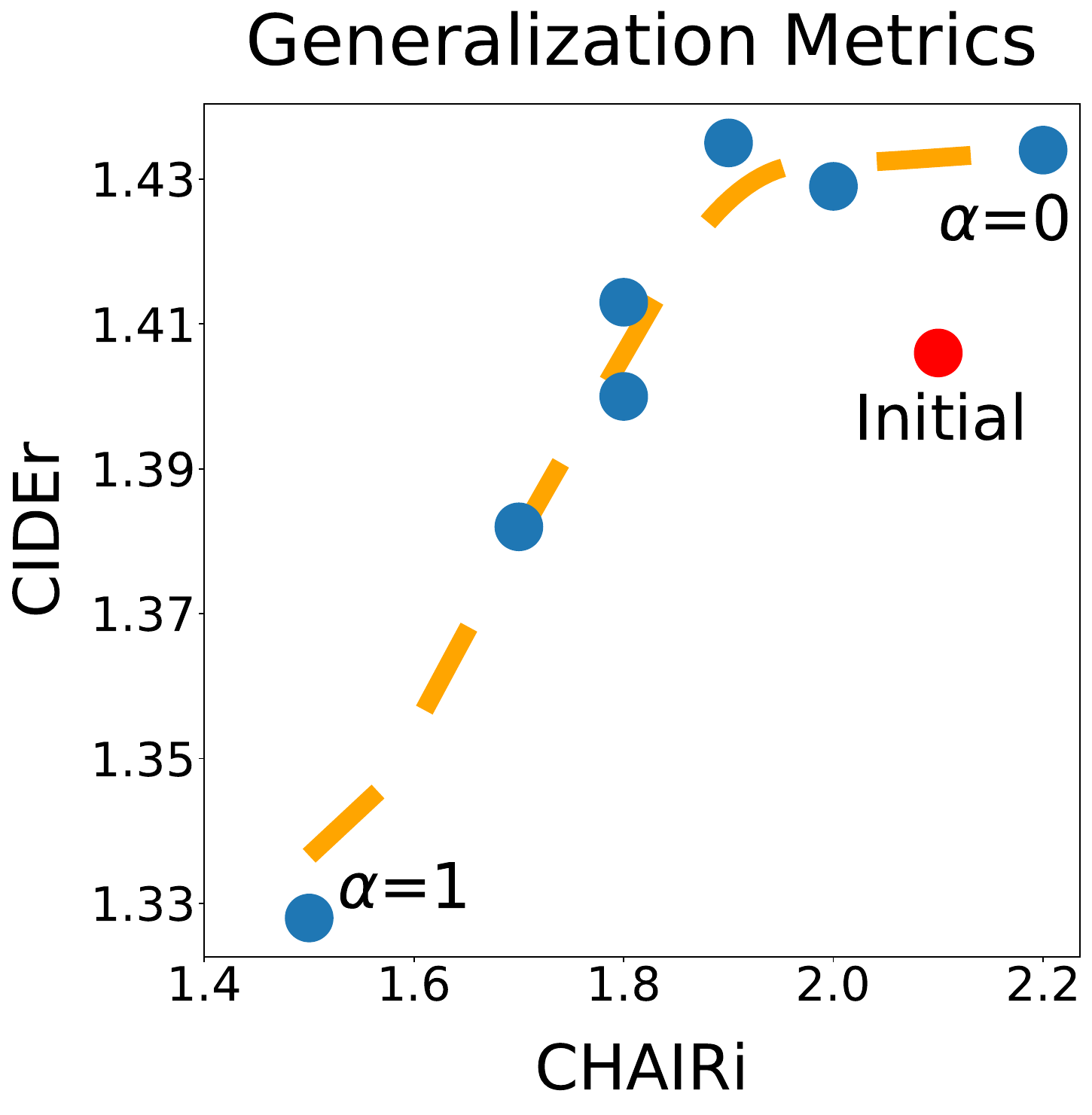}}{}
    \caption{\textbf{Fidelity-Adequacy graphs} for pretrained (``initial'') and \methodname{}-optimized BLIP models. As seen above, varying the reward weighting $\alpha$ adjusts the trade-off between caption fidelity (x-axis) and adequacy (y-axis), with intermediate values outperforming the initial model (``Initial''). This holds both for metrics we directly optimize (left) and additional metrics (right), illustrating the generalization ability of our approach.
    }
    \vspace{-0.1in}
    \label{fig:fidelity_adequacy}
\end{figure}

%% file: tables/mocha_comparisons.tex
\begin{table}[t]
  \centering
  \setlength{\tabcolsep}{1pt}
    \begin{tabular}{lcccccccc}
      \toprule
      &\multicolumn{2}{c}{\multirow{2}{*}{\shortstack{Quality}}} & \multicolumn{4}{c}{Hallucination}\\
      &  & &\multicolumn{2}{c}{Closed} & \multicolumn{2}{c}{Open} \\
      Model& B@4$\uparrow$ & C$\uparrow$ 
      & {CH$_i$}$\downarrow$ & {CH$_s$}$\downarrow$ & OCH $\downarrow$ & $\bar{p}\downarrow$ & \\ %

        \midrule

        BLIP & 41.5 & 138.4 & 2.3 & 3.5 & 19.2 & 0.244 \\ %

        BLIP+L  & 5.5 & 0.0 & 12.1 & 35.4 & 31.8 & 0.321 \\ %
        
        BLIP+T  & 41.3 & 137.4 & \textbf{1.9} & \textbf{2.8} & 19.2 & 0.241 \\ %
        
        BLIP+M & \textbf{41.9} & \textbf{139.6} &  2.1 & 3.1 & \textbf{18.3} & \textbf{0.206} \\ %
        \midrule

        BLIP-2 & 43.4  &	\textbf{144.3} & 	1.7 & 	2.6 & 	17.0  &  0.207 \\ %

        BLIP-2+L  & 5.7 	& 0.0 	& 12.1 & 	33.6 & 	28.4 & 0.259 \\ 
        
        BLIP-2+T  & 43.3 & 	143.5 	& \textbf{1.3} 	& \textbf{2.0} 	& 17.0 & 0.206  \\ 
        
        BLIP-2+M & \textbf{44.0} 	& \textbf{144.3} 	& 1.4  & 2.3 & \textbf{16.6} &  \textbf{0.199} \\
        
        \bottomrule

    \end{tabular}
  \vspace{-5pt}
  \caption{\textbf{Comparison To Prior Works}. Measured for BLIP-Large and BLIP-2. +L/T/M refer to LURE, TLC-A, and \methodname{} respectively. B@4, C, CH, OCH, and $\overline{p}$ denote BLEU-4, CIDEr, CHAIR, OpenCHAIR, and NLI $p(\text{contr.})$ metrics respectively. All metrics are measured over MS-COCO test set, except for OCH which is measured over our OpenCHAIR benchmark. %
  }
  \vspace{-0.26in}
  \label{tab:comparisons_mocha}
\end{table}

%% file: 02-rw.tex
\input{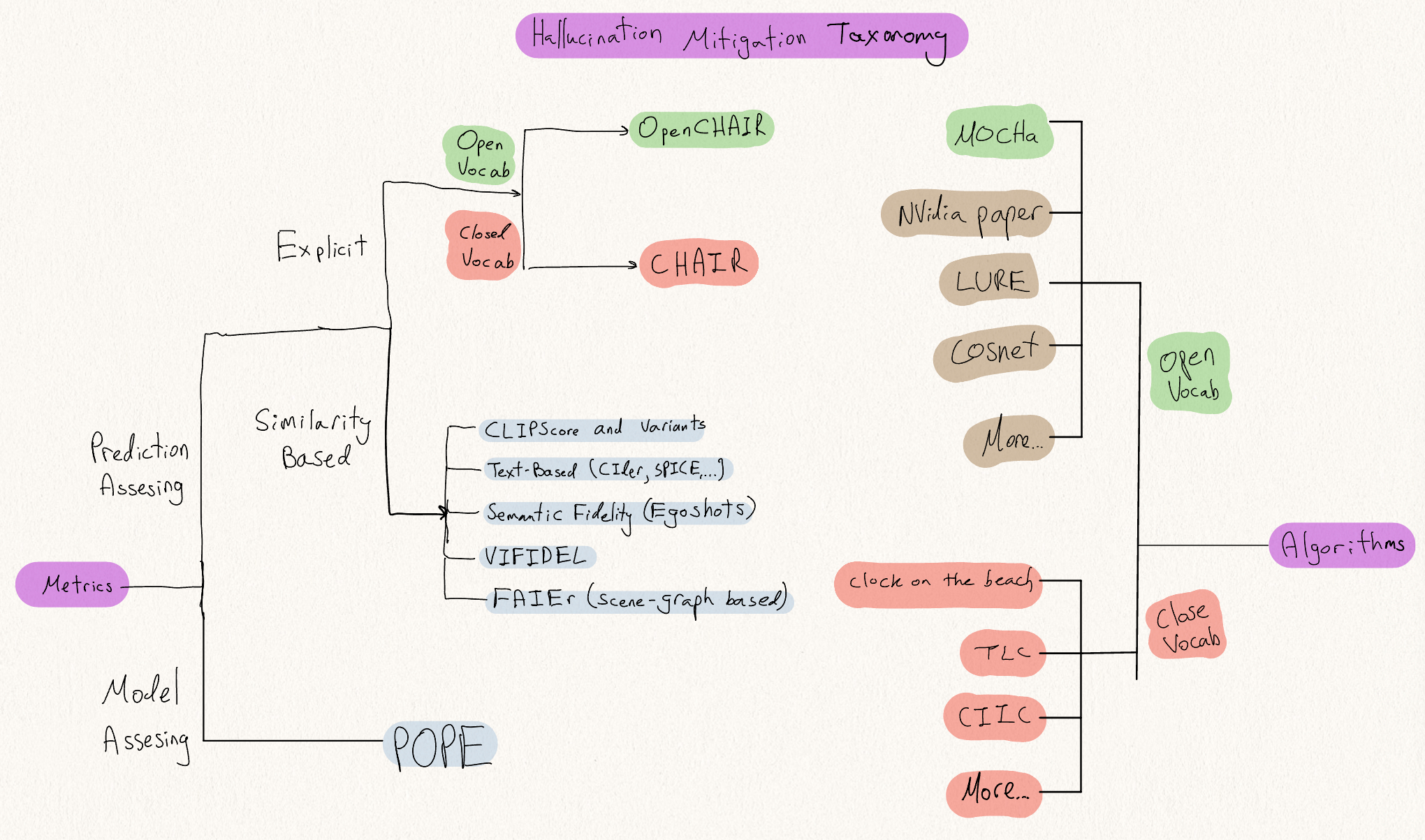}

\section{Related Work} \label{sec:rw}
We provide a short summary of related works here, with an extended discussion of their methods and differences from our work in the appendix.

\noindent \textbf{Measuring VLM Hallucinations.}
Several works have proposed holistic measures of generated text fidelity with respect to an input image using embedding similarities or learned metrics; such methods (the ``Similarity Based'' metrics of Figure \ref{fig:taxonomy}) include CLIPScore and variants~\cite{hessel2022clipscore, shi2022emscore}, Semantic Fidelity~\cite{agarwal2020egoshots}, VIFIDEL~\cite{madhyastha-etal-2019-vifidel}, and FAIer~\cite{wang2021faier}. 
While these metrics may correlate with the presence of hallucinations, they are less interpretable as they do not provide a discrete count of hallucinations in a predicted caption.
By contrast, the POPE metric~\cite{li2023pope} compares ground-truth objects with a model's answers when asked if each object is present; this is open-vocabulary but differs from our setting as it does not score predicted captions but rather assesses a VQA model's general knowledge (indicated as ``Model Assessing'' in Figure \ref{fig:taxonomy} (left)).

\noindent \textbf{Reducing VLM Hallucinations.}
Various methods for mitigating hallucinations in image captioning have been proposed (see Figure \ref{fig:taxonomy} (right)). Until recently, research on mitigating hallucinations in captions has largely considered object (noun) hallucinations, typically confined to a closed vocabulary, for instance, objects defined in MS-COCO. Such works include UD-L~\cite{biten2021clockonthebeach}, CIIC~\cite{liu2022ciic}, TLC~\cite{petryk2023tlc}, ObjMLM~\cite{Dai2023Plausible}, and Woodpecker~\cite{yin2023woodpecker}. 
Unlike these works, we mitigate hallucinations in the more challenging open-vocabulary setting. The contemporary work LURE~\cite{Zhou2024Analyzing} proposes a method for the open setting, but their proposed approach (complementary to ours) was not evaluated automatically in an open vocabulary setting due to the lack of an existing benchmark. Figure \ref{fig:taxonomy} illustrates which explicit hallucination metric was used to evaluate each algorithm.

As instruction-following VLMs rapidly develop, multiple concurrent works have considered hallucinations in related tasks such as visual question-answering (VQA), applying RL-based methods adopted from research on LLMs~\cite{gunjal2023detecting,sun2023llavarlhf,sun2023salmon}.
These methods, which do not directly target our task, also require laborious human annotation to train a supervised reward model to penalize hallucinations, while our approach does not require any explicit supervision.

\noindent \textbf{Deep RL for VLM Text Generation.} 
Deep RL has been widely applied to text generation tasks and specifically for optimizing classical image-captioning metrics~\cite{rennie2017self,stefanini2022show}.
Another more recent development is the rise of deep RL for LLMs, which commonly uses the Reinforcement Learning from Human Feedback (RLHF) framework, which requires manual human preference annotation for training a reward model~\cite{ziegler2020finetuning,stiennon2020learning,ouyang2022training}.
Beyond LLMs, RLHF has been recently applied to aligning multimodal models with human preferences~\cite{abramson2022improving}. While such methods succeed in optimizing sequence-level properties, they often suffer from increased hallucinations as a side-effect of optimizing for human preferences or standard NLG sequence-level metrics (as illustrated in Appendix~\ref{sec:add_comp}).

%% file: figures/taxonomy/taxonomy.tex
\begin{figure*}
    \centering
    \includegraphics[trim={2.1cm 6.1cm 1.5cm 2cm}, clip, width=0.95\linewidth]{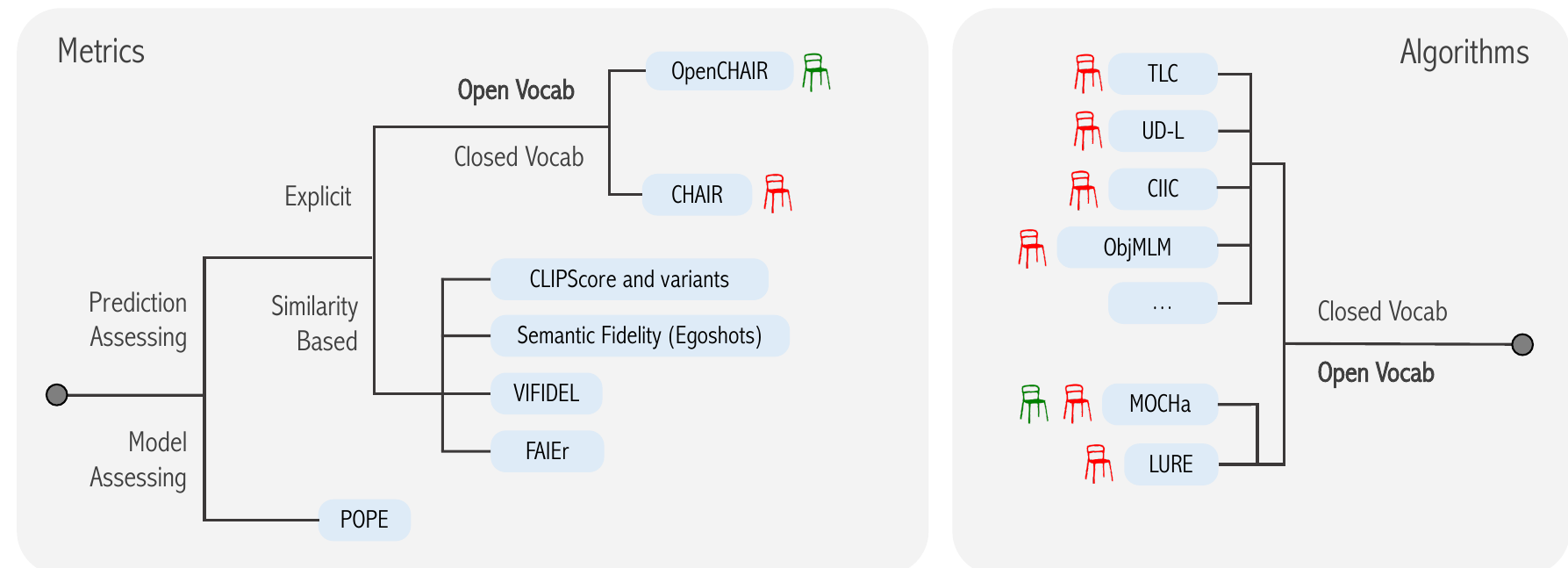}
    \vspace{-2pt}
    \caption{\textbf{VLM Caption Hallucination Taxonomy}. We illustrate metrics (left) and algorithms (right) for quantifying and mitigating hallucinations in image-conditioned text generation. We propose an explicit metric for measuring open-vocabulary hallucinations (\metricname{}) and an open-vocabulary hallucination mitigation algorithm (\methodname{}).
    We mark each algorithm with the automatic hallucination rate metric with which it is evaluated ({\color{nicegreen}Green} -- \metricname{}, {\color{red}Red} -- CHAIR).
    Further details
    are provided in Section \ref{sec:rw}.
    }
    \label{fig:taxonomy}
    \vspace{-0.2in}
\end{figure*}

%% file: 06-conclusion.tex
\section{Conclusion}
We have shown the significance of operating in an open-vocabulary setting to effectively quantify and mitigate caption hallucinations. These are explicitly measured by our \metricname{} benchmark, and our \methodname{} framework allows for optimizing captioning models to reduce such hallucinations while preserving caption quality. This reduction is demonstrated on our benchmark and other existing metrics. Our method and benchmark may be applied flexibly to a variety of model sizes and architectures, which we foresee providing a framework for future work on hallucination-aware captioning.

\section{Limitations}

While \metricname{} provides diverse coverage of object types, it does not directly measure non-object hallucinations (e.g. hallucinated attributes or relations between entities), which are also targeted by sequence-level approaches such as our \methodname{} optimization. We have focused on objects as a natural extension of the existing closed-vocabulary object hallucination benchmark CHAIR, and due to the fact that extracting and comparing objects from image captions is a relatively well-defined task. Future work may consider extending our OpenCHAIR concept to non-objects, specifically, constructing a robust benchmark for evaluating hallucinations on the attribute-, relation-, predicate-level, or of other types, utilizing elements of our methodology such as open-vocabulary LLM evaluation. Furthermore, we acknowledge that captioning models may show different performance on the synthetic images found in \metricname{} relative to natural images, although we have found it to correlate empirically to other hallucinations metrics and human intuition.

We emphasize that our work does not solve the hallucination problem completely, although it presents a significant step towards this goal. Note also that we have focused in this work on the image captioning domain, while modern VLMs are often applied to diverse tasks such as VQA and visual instruction-following for which hallucinations also pose a significant challenge. We hope that our proposed strategy will pave the way for future research on hallucination reduction in all of these domains, in which open-vocabulary approaches also present significant promise.

%% file: 07-impact-statement.tex
\section{Ethics Statement}
\label{sec:broader_impact}
\vspace{-8pt}
This work focuses on measuring and mitigating hallucinations in visual-language models (VLMs). As such it is expected to increase the reliability of VLMs and the ability to measure their performance, which is important when using them in real world systems. This is expected to have a positive impact on the use of VLMs in the society. However, we do recognize that the foundation models used in the \metricname{} construction and evaluation pipeline and those used to calculate the \methodname{} reward function could propagate biases. We anticipate further research into such biases before relying on our work beyond the research environment.

%% file: supp_00_viz.tex
\clearpage
\section{Interactive Visualization}

For additional qualitative results, we refer the reader to the interactive visualization tool provided at \url{https://assafbk.github.io/mocha_vis_tool}.

We provide image captioning results using BLIP-Large with and without \methodname{} for 350 randomly selected test images from MS-COCO~\cite{lin2015mscoco} and Flickr30K~\cite{young2014flickr30k}. %

To visually emphasize the hallucination rate in the predictions, for each model we calculate the NLI contradiction probability\footnote{Using the same pretrained NLI model described in the main paper.} between the top beam and a ground-truth caption (which is depicted below the image), and report the difference in the contradiction probability between the two models. Samples are ordered via n-gram similarity between the predictions of both models, listing the most different predictions first, allowing for better emphasizing items with evident differences first. This is calculated by considering the top 5 beams of BLIP as reference texts and the top 5 beams of BLIP+\methodname{} as candidate sentences; we then compute the average BLEU~\cite{papineni2002bleu} score between each candidate and all references.

%% file: supp_01_details.tex
\section{Additional Details}

\subsection{\methodname{} Implementation Details}

As discussed in Rennie et al.~\cite{rennie2017self}, we reduce variance in gradient estimates by shifting the reward function to have zero mean; we apply this to the reward function before adding the KL penalty. We achieve this by subtracting the sample mean of this reward (without KL penalty) from all predictions for a given image in a minibatch. 

During each training iteration, we build minibatches by selecting 10 images and then generating 10 predictions per image (hence 100 image-prediction pairs total). We use nucleus sampling~\cite{holtzman2019curious} with $p=0.9$ and temperature $t = 1.2$, and we cap generations to be at most 40 tokens. We apply PPO with clipping parameter $\epsilon=0.2$. For our reward function, we use coefficients $\alpha=0.5$ and $\beta \in [0.004, 0.06]$ (depending on the model optimized).

During \methodname{} training, we freeze the image encoder of all models, training the text encoder components alone. For BLIP-Large and BLIP-Base we use gradient clipping of 5, learning rate of 1e-6 and 4 PPO steps in each iteration. BLIP-2 is trained with low rank adapters (LoRA) over the keys and values of the decoder attention layers~\cite{hu2021lora} with a learning rate of 1e-6. GIT-base is trained with a learning rate of 1e-5 with 4 PPO steps and gradient clipping of 5.

All model checkpoints are taken from the Hugging Face Model Hub\footnote{https://www.huggingface.co/models}):
\begin{itemize}
    \setlength{\itemindent}{-.15in}
    \item \texttt{salesforce/blip-image-captioning-large}
    \item \texttt{salesforce/blip-image-captioning-base}
    \item \texttt{salesforce/blip2-opt-2.7b-coco}
    \item \texttt{microsoft/git-base-coco}
\end{itemize}
We train these models for the following number of iterations: 350 for BLIP-B, 1200 for BLIP-L, 3400 for BLIP-2, and 600 for GIT-B.

\subsection{\metricname{} Implementation details} \label{sec:och_details_supp}

\input{tables/openChair_ablation_supp}
\noindent \textbf{Generating Diverse Captions}
\noindent \quad
We start by parsing all objects in MS-COCO's human-annotated captions by first identifying nouns via syntactic parsing\footnote{Using the $en\_core\_web\_md$ pipeline from the SpaCy~\cite{spacy2} library.}. We then filter these for highly concrete nouns, by using the values recorded by Hessel et al.~\cite{hessel2018quantifying} with threshold 4.5. We used these objects, coupled with their corresponding captions, to prompt an instruction-tuned LLM\footnote{\label{footnote:llama_70b}\texttt{meta-llama/Llama-2-70b-chat-hf (4-bit quant.)}} to rephrase the captions with different objects. We used stochastic sampling with top-p of 0.9 and temperature of 0.6 for this LLM generation. While this stage increases the object diversity, we notice that the output still includes many common objects that have a significant overlap with those in MS-COCO. To overcome this issue, we filter out all captions that do not include rare objects, defining an object as rare if its appearance frequency in the dataset is in the lowest 10th percentile.
The remaining captions are used as few-shot examples for a LLM\footnote{\texttt{meta-llama/Llama-2-13b}} (base, not instruction-tuned) to generate new captions, to further increase diversity. We used 10 few shot example for each generated caption, and text is generated using sampling  with temperature 0.8.  We generate 5,000 captions from the LLM and feed them as prompts to the text-to-image model Stable Diffusion XL~\cite{podell2023sdxl}, which generates a single image for each caption. For image generation, we use 40 sampling steps and guidance scale of 10. We also employ negative prompting using the prompt \emph{``unclear, deformed, out of image, disfigured, body out of frame"} to encourage generation of clear objects in the output images.

\medskip
\noindent
\textbf{Evaluation on the \metricname{} Benchmark} 
\noindent \quad
Evaluating a captioning model on \metricname{} is performed as follows: First, all the objects in the caption generated by the captioning model are extracted using the parsing method described in the previous paragraph. For each detected object, an LLM\footnoteref{footnote:llama_70b} is prompted to determine whether the object is in the GT caption or not using the prompt: \emph{``<s>[INST] An image has the following caption: ``$\langle$input caption$\rangle$".
Does the image contain the following object? ``$\langle$input object$\rangle$".
Answer yes/no/unsure.
The answer is: [/INST]"} . We use greedy decoding for this stage. Objects for which the LLM answers ``no'' are counted as hallucinations and objects for which the LLM answers ``yes'' are counted as existing objects. We ignore objects that receive any other response, and report that the amount of such objects are <2\% of the total objects considered. Finally, the \metricname{} hallucination rate is calculated as $OCH := n_h / (n_{h}+n_{e})$, where $n_h$ is the number of hallucinated objects and $n_e$ is the number of existing objects. We note that we added a short list of objects to ignore: \textit{['painting', 'drawing', 'photo', 'picture', 'portrait', 'photograph']}. Since the prefix of the prediction tends to have the following form: \textit{``A photograph of...'', ``A picture of...''}, these words are identified as concrete objects and then classified as hallucinations by the LLM (as they dont appear in the GT caption), hence should be ignored.

\input{figures/precision_recall_TLC/precision_recall_TLC}
\input{figures/qualitative/kl_ablation}

\input{tables/correlation_och_syn}
\input{tables/correlation_ch_coco}

\subsection{LURE Comparison}
\label{sec:lure-details}

To evaluate LURE~\cite{Zhou2024Analyzing} in our setup, we followed the authors' instructions\footnote{\url{https://github.com/YiyangZhou/LURE/blob/main/README.md}}  and applied their pre-trained model (\texttt{YiyangAiLab/LURE}, over MiniGPT-4 with VICUNA-13b) to our predicted captions. Both BLIP-L's and BLIP-2's predictions (with beam search decoding, 5 beams) were supplied to LURE's revisor along with the probabilities of each predicted token for the highest scoring beam. After additional parsing, LURE revised BLIP-L's and BLIP-2's sentences, which we then evaluated with various metrics. An example of this procedure is given below:

\begin{itemize}
\item Original caption from BLIP-L, masked by LURE's algorithm: \emph{a woman standing in a kitchen with [IDK]}
\item LURE instruction: \emph{According to the picture, remove the information that does not exist in the following description:}
\item LURE output revision: \emph{In the image, a smiling woman is standing in a kitchen with various food items spread out on a table. She seems to be posing and enjoying the scene, perhaps taken during a cooking session. The table is filled with food, including several pieces of sushi placed near the center, along with some pasta on the left side, cucumbers near the right side, and a couple of apples towards the back.}
\end{itemize}

\subsection{TLC-A Comparison}
\label{sec:tlc-details}
In order to compare our method to TLC-A~\cite{petryk2023tlc}, we received code from its authors and implemented it in our setup. TLC-A is a decoding-time method applied to auto-regressive captioning models, and in our setting we apply it to different models (e.g. BLIP-Large) than those tested by Petryk et al (e.g. OFA). Of particular note is that TLC-A requires selecting a threshold confidence value, which is used in the decoding phase to re-rank generated beams according to the confidence assigned to COCO object tokens. Petryk et al. recommend calibrating this threshold using the COCO validation set to achieve a precision level of at least 99\%; however, in our experiments we find that this value cannot be achieved by the models we consider without sacrificing most of the recall, as illustrated in Figure \ref{fig:pr_tlc}.
Therefore, we instead use the COCO validation set to select the best-performing threshold with respect to the CHAIR metric, as shown in Table \ref{tab:tlc_th_table}. The selected confidence threshold is 0.33 and it achieves a precision of 98.3\% and a recall of 84\% over the validation set.

\input{tables/tlc_th_table}

%% file: figures/precision_recall_TLC/precision_recall_TLC.tex
\begin{figure}[ht]
    \centering
    
    \includegraphics[width=1\linewidth]{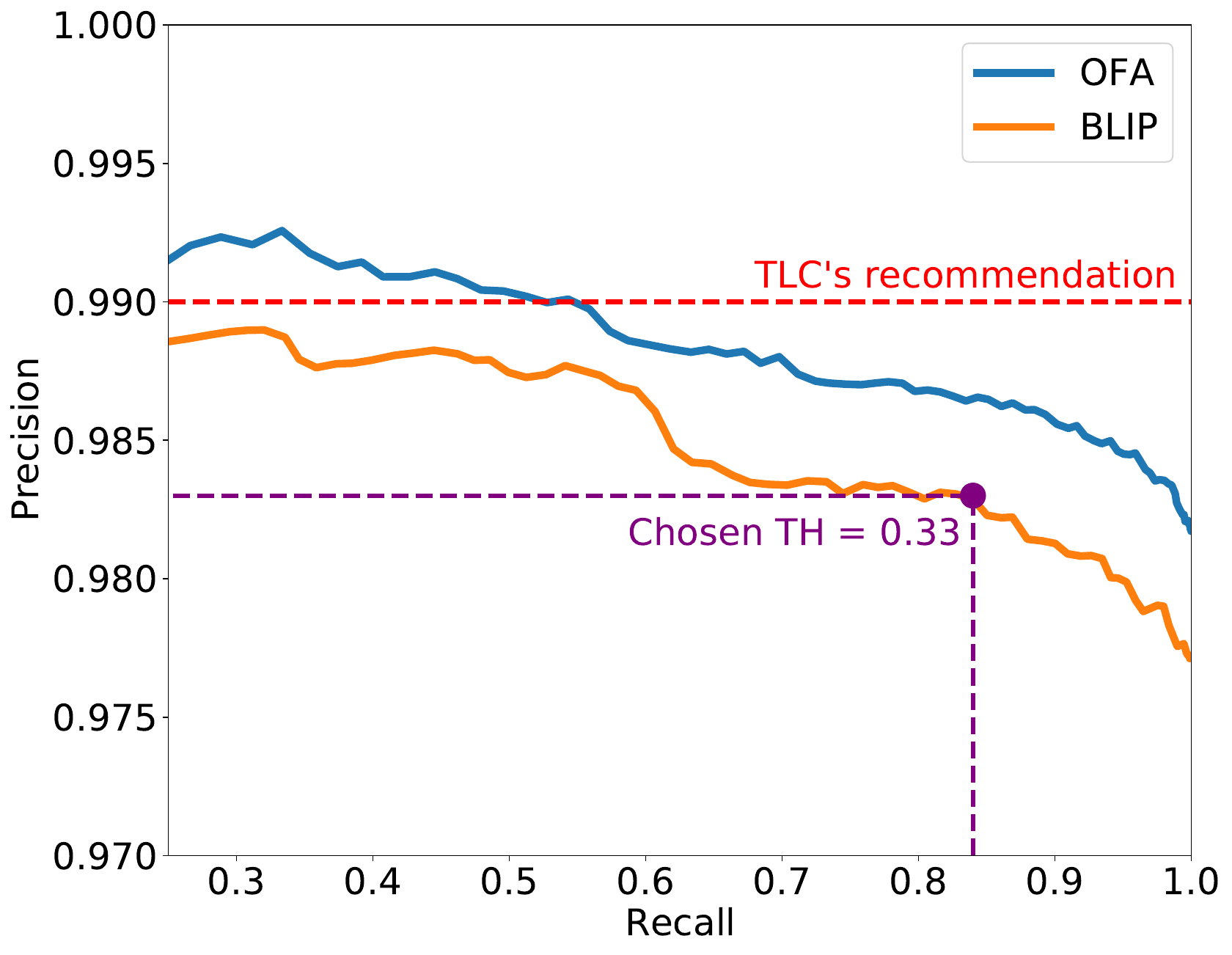}
    \caption{\textbf{Precision-recall curve for selecting TLC-A threshold.} As detailed in ~\cite{petryk2023tlc}, we compute a precision-recall curve over the predicted object confidences. As illustrated above, the 99\% precision threshold recommended by Petryk et al.~\cite{petryk2023tlc} cannot be achieved by BLIP-Large on the COCO Karpathy validation set. Hence, in our setting we must adjust the threshold to find a reasonable balance between precision and recall.}
    \label{fig:pr_tlc}
\end{figure}

%% file: figures/qualitative/kl_ablation.tex
\begin{figure}[ht]
    \begin{minipage}{\linewidth}
        \begin{tabularx}{\linewidth}[t]{|s|R|R|}
            \hline
            \multicolumn{1}{|c|}{\raisebox{0.8\height}{\includegraphics[width=0.07\linewidth]{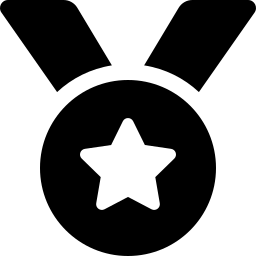}}}
            & \lapbox[\width]{-0.5em}{\raisebox{-1.12mm}{\includegraphics[trim={0pt 0pt 0pt 0pt},clip,height=2.6cm]{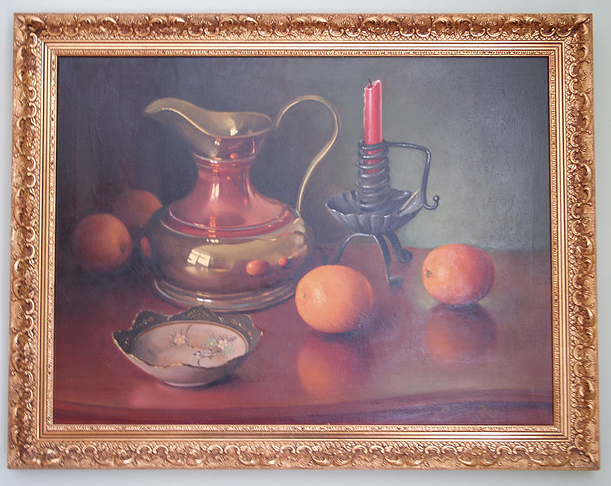}}}
            & \lapbox[\width]{-0.4em}{\raisebox{-1.14mm}{\includegraphics[trim={45pt 0pt 20pt 0pt},clip,height=2.6cm]{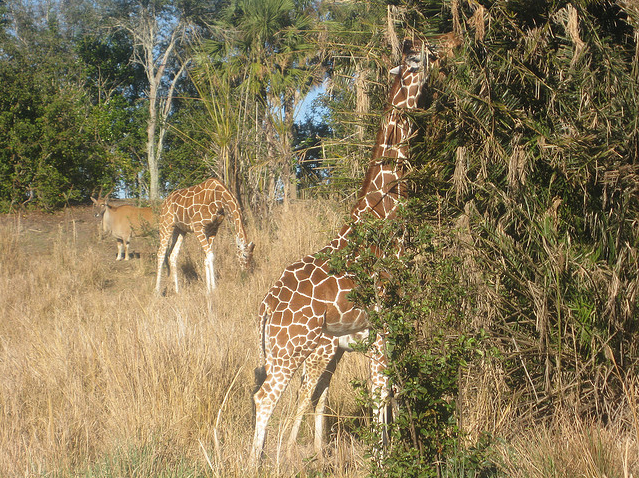}}}
            \\
            \hline
            \multicolumn{1}{|c|}{\raisebox{-0.5\height}{$\emptyset$}}
            & \small{\emph{a painting of oranges and a silver pitcher on a table}}
            & \small{\emph{two giraffes eating leaves from a tree}} \\ \hline
            
            \multicolumn{1}{|c|}{\raisebox{-0.5\height}{$-r_{kl}$}}
            & \small{\emph{a painting of some items}}
            & \small{\emph{some giraffes in the field}} \\ \hline

            \multicolumn{1}{|c|}{\raisebox{-0.5\height}{$r$}}
            & \small{\emph{a painting of a pitcher, oranges, and a candle on a table}}
            & \small{\emph{a giraffe eating leaves from a tree in a field}} \\ \hline
        \end{tabularx}
    \end{minipage}
    \vspace{-4.3pt}
    \caption{\textbf{Ablating the KL-penalty reward}. Above we show captions sampled from various models: the initial model (BLIP-Large) before optimization ($\emptyset$), the model with \methodname{} optimization applied and KL penalty ablated ($-r_{kl}$), and an optimized model with our full reward function ($r$). As is seen above, while the base model outputs various hallucinations (e.g. \emph{a silver pitcher}), the model optimized without KL penalty outputs generic texts without adequate detail, due to over-optimization of the fidelity objective. Optimizing with the full reward function yields captions that are both descriptive and consistent with the input condition.}
    
    \label{fig:qual_kl_reward_abl}
\end{figure}

\footnotetext{Reference ground truth captions: \emph{Painting of oranges, a bowl, candle, and a pitcher} (left) and \emph{A giraffe grazing on a tree in the wilderness with other wildlife} (right).}

%% file: tables/correlation_och_syn.tex
\begin{table}[t]
  \centering
  \setlength{\tabcolsep}{2pt}
    \begin{tabular}{l|cc}
        \textbf{BLIP2}           & Pred = `E' & Pred = `H' \\
        
        \midrule
        
        GT = `E'        & 332       & 42  \\

        GT = `H'        & 0         & 54 \\ 
       
    \end{tabular}
       
    \vspace{3mm}
    \begin{tabular}{l|cc}
    
        \textbf{BLIP-L}           & Pred = `E' & Pred = `H' \\
        
        \midrule
        
        GT = `E'        & 353       & 44  \\

        GT = `H'        & 0         & 31 \\

    \end{tabular}

    \vspace{3mm}
    \begin{tabular}{l|cc}
    
        \textbf{GIT-B}           & Pred = `E' & Pred = `H' \\
        
        \midrule
        
        GT = `E'        & 325       & 36  \\

        GT = `H'        & 1         & 66 \\

    \end{tabular}

    \vspace{3mm}
    \begin{tabular}{l|cc}
    
        \textbf{OFA-L}           & Pred = `E' & Pred = `H' \\
        
        \midrule
        
        GT = `E'        & 336       & 45  \\

        GT = `H'        & 1         & 46 \\

    \end{tabular}

  \caption{\textbf{Human Evaluation of \metricname{} Benchmark}. The tables illustrate a correlation measurement between \metricname{}'s automatic hallucination annotations (Pred) and manual human hallucination annotations (GT). `E', `H' stand for 'object \textbf{E}xists', 'object \textbf{H}allucinated', respectively.  BLIP2, BLIP-L, GIT-B and OFA-L stand for BLIP2-2.7b, BLIP-Large, GIT-Base, OFA-Large, all fine-tuned for image-captioning over COCO.}
  \label{tab:corr_och_syn}
\end{table}

%% file: tables/correlation_ch_coco.tex
\begin{table}[t]
  \centering
  \setlength{\tabcolsep}{2pt}

    \begin{tabular}{l|cc}
        \textbf{BLIP2}           & Pred = `E' & Pred = `H' \\
        
        \midrule
        
        GT = `E'        & 416       & 3  \\

        GT = `H'        & 4         & 5 \\ 
       
    \end{tabular}
       
    \vspace{3mm}
    \begin{tabular}{l|cc}
    
        \textbf{BLIP-L}           & Pred = `E' & Pred = `H' \\
        
        \midrule
        
        GT = `E'        & 413       & 2  \\

        GT = `H'        & 4         & 9 \\

    \end{tabular}

    \vspace{3mm}
    \begin{tabular}{l|cc}
    
        \textbf{GIT-B}           & Pred = `E' & Pred = `H' \\
        
        \midrule
        
        GT = `E'        & 412       & 1  \\

        GT = `H'        & 3         & 12 \\

    \end{tabular}

    \vspace{3mm}
    \begin{tabular}{l|cc}
    
        \textbf{OFA-L}           & Pred = `E' & Pred = `H' \\
        
        \midrule
        
        GT = `E'        & 418       & 2  \\

        GT = `H'        & 3         & 5 \\

    \end{tabular}

  \caption{\textbf{Human Evaluation of CHAIR Benchmark}. The tables illustrates a correlation measurement between CHAIR's automatic hallucination annotations (Pred) and manual human hallucination annotations (GT). `E', `H' stand for 'object \textbf{E}xists', 'object \textbf{H}allucinated', respectively. BLIP2, BLIP-L, GIT-B and OFA-L stand for BLIP2-2.7b, BLIP-Large, GIT-Base, OFA-Large, all fine-tuned for image-captioning over COCO.}
  \label{tab:corr_ch_coco}
\end{table}

%% file: tables/tlc_th_table.tex
\begin{table*}[t]
  \centering
  \setlength{\tabcolsep}{2pt}
    \begin{tabular}{c|cc|ccccccccccc}
        \toprule
        TH & P & R & B@4$\uparrow$ & C$\uparrow$ & {CH$_i$}$\downarrow$ & {CH$_s$}$\downarrow$ &
        $\bar{p}\downarrow$ & BSc $\uparrow$ &  \\
        
        \midrule
        -     & -     & -     & 41.5 & 138.4 & 2.3 & 3.5 & 0.246 & 0.679 \\
        \midrule
        0.10 & 0.978 & 0.99 & 41.4 & 138.0 & 2.2 & 3.38 & 0.246 & 0.677  \\
        0.21 & 0.980 & 0.94 & 41.4 & 137.7 & 2.1 & 3.14 & 0.243 & 0.677  \\
        \textbf{0.33} & \textbf{0.983} & \textbf{0.84} & \textbf{41.2} & \textbf{137.5} & \textbf{1.91} & \textbf{2.82} & \textbf{0.243} & \textbf{0.676}  \\
        0.52 & 0.986 & 0.61 & 41.1 & 136.7 & 1.97 & 2.9 & 0.242 & 0.675  \\
        0.56 & 0.988 & 0.55 & 41.2 & 136.8 & 1.94 & 2.86 & 0.243 & 0.675  \\
        0.94 & 1     & 0.01 & 41.4 & 137.7 & 2.21 & 3.32 & 0.247 & 0.677  \\

        \bottomrule
    \end{tabular}

  \caption{\textbf{Selecting a threshold for TLC-A.} We evaluate TLC-A with different thresholds (as described by Petryk et al.~\cite{petryk2023tlc}) over the COCO caption Karpathy validation set. In the first row we have BLIP without TLC-A. We indicate the selected threshold which achieves the best CHAIR scores overall \textbf{in bold}. B@4, C, CHi, CHs, BSc, $\overline{p}$ denote BLEU-4, CIDEr, CHAIR instance and CHAIR sentence, BERTScore, and NLI $p(\text{contr.})$ metrics respectively. P, R are the precision and recall that each threshold (for predicted object confidences) achieves over the validation set.
  } 
  \label{tab:tlc_th_table}
\end{table*}

%% file: supp_02_comparisons.tex
\input{arxiv/tables/results_open_vocab_arxiv}

\input{tables/additional_ablation_supp}

\section{Additional Results}
\label{supp_additional_comparisons_and_ablations} \label{sec:supp_abl}

\subsection{Full Quantitative Results}
We show in Table \ref{tab:full_sota_table} the full results, comparing the \methodname{} optimized models (marked by +M) to the baselines (Figure \ref{fig:main_table} was prepared using this data). Since there is only about a 1 point improvement in OCH, it may seem that \methodname{} does not generalize well to other datasets.
First, to alleviate the concern regarding the relatively small improvement in OCH scores after \methodname{}-tuning, we perform an additional analysis to interpret this result. Focusing on images that are prone to object hallucination (where the model predicts a hallucinated object either before or after \methodname{}-tuning, occurring on approximately half of the images), we observe, e.g., a 7 point improvement in OCH score (20\% improvement) for Blip-Large. We conclude that \methodname{} does improve hallucinations, especially at the hard examples, yet it is harder to perform this distinction since there are many 'easy' objects that are detected by both models (before and after fine-tuning), which thanks to the 'precision' nature of the calculation decreases the improvement in OCH score. The 'hard question' improvements for the other models are: 10 points for Blip-Base, 4 points for Blip-2 and 13 points for Git-Base.
Second, in Appendix \ref{appendix:generalization} we show that \methodname{} generalizes to additional datasets and styles.

\input{tables/results_closed_vocab}

\subsection{Comparisons of \metricname{} and CHAIR}
\label{appendix_comparison_och_ch}

In Tables \ref{tab:corr_och_syn}--\ref{tab:corr_ch_coco} we provide full numeric results for our human evaluation of \metricname{} and CHAIR across a variety of captioning model predictions, as we discuss in the main paper.

In Figure \ref{fig:och_cov}, we illustrate the number of unique object types found in these benchmarks. We note that \metricname{} contains a much larger diversity of object types, even when considering the full contents of CHAIR's synonym list.

\input{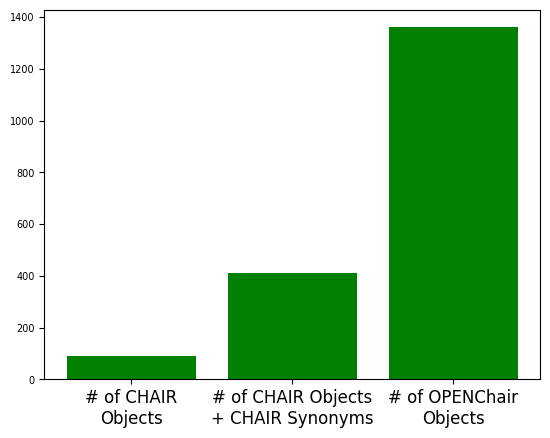} 

\subsection{\metricname{} Efficiency}
\label{appendx_och_efficiency}

We calculate \metricname{} using Llama-3-8B-Instruct and find that the evaluation and overall improvement trends are similar to Llama-2-70B-chat-hf (4-bit quant.). This can be seen in Table \ref{tab:efficient_och}, showing improvements in OCH scores after MOCHa-tuning.

\input{tables/efficient_och}

\subsection{Additional Ablations} \label{sec:supp_additional_abl}

\input{tables/results_abl}

\input{figures/qualitative/reward_ablation_figure_multiimage}

\noindent \textbf{Reward Ablations.}
In Table \ref{tab:abl_table}, we provide numeric results for ablating the fidelity and adequacy terms in our reward function. As discussed in the main paper, removing either of these reward terms leads to a degradation with respect to either hallucinations or textual quality, while using both together displays a synergistic effect with hallucinations reduced (as reflected by metrics such as CHAIR) while preserving or even improving caption quality (as reflected by general textual quality metrics such as BLEU-4). We also show a qualitative illustration of these results in Figure \ref{fig:qual_reward_abl}.

We demonstrate the effect of our KL penalty in the reward function by performing \methodname{} optimization without this term. As can be observed in the fifth row of Table \ref{tab:additional_ablations_table_supp}, optimization without this penalty improves the NLI-based reward $\bar{p}$ while degrading other measures of text quality (including non-optimized metrics like CIDEr). We hypothesize that allowing the model to freely deviate from its initial distribution encourages it towards a degenerate solution with respect to $\bar{p}$, which may be the easiest reward term to over-optimize in an unconstrained setting. This is also reflected qualitatively as seen in Figure \ref{fig:qual_kl_reward_abl}. As illustrated in the figure, captions generated by the model trained without the KL penalty ($-r_{kl}$) do not contradict the image, but rather contain generic text (e.g. \emph{a painting with some items}), lacking adequate detail. By contrast, optimizing with the KL penalty reward yields captions that are both descriptive and consistent with the input condition, reflected in the improved scores across metrics in Table \ref{tab:additional_ablations_table_supp} and the quality of predictions of the full reward model ($r$) in Figure \ref{fig:qual_kl_reward_abl}. This is attributed to the ability of the KL penalty to mitigate over-optimization, which benefits both optimized rewards.

\medskip
\noindent
\textbf{PPO Ablation.}
We also ablated the selection of RL algorithm, by replacing PPO with the SCST algorithm upon which it is based (noting that SCST is the common name for the REINFORCE algorithm in the context of image captioning)~\cite{Sutton1998,schulman2017ppo,rennie2017self}. As is seen in Table \ref{tab:additional_ablations_table_supp}, PPO outperforms SCST across metrics, consistent with prior work on PPO finding that it avoids instabilities during optimization that may allow it to converge to a more optimal solution~\cite{schulman2017ppo, ouyang2022training, ziegler2020finetuning}.

\subsection{Additional Comparisons}
\label{sec:add_comp}

\textbf{Comparison to Dedicated Models}
In Table \ref{tab:closed_vocab} we provide full numeric results for older dedicated models compared to a modern VLM without further optimization, showing that they are outperformed by all metrics.

\noindent
\textbf{Comparison to RLHF-Tuned VLMs.}
LLaVa-RLHF \cite{sun2023llavarlhf} is a concurrent work, which aims to reduce hallucinations in instruction tuned models using factually-grounded RLHF. In Table \ref{tab:llava_rlhf_comp}, we provide a quantitative comparison between LLaVa-RLHF and BLIP+\methodname{} over 100 samples of the OPENChair benchmark. For LLaVa-RLHF decoding we use both stochastic sampling with the default parameters recommended by the authors, as well as greedy sampling (as beam search is not implemented for LLaVa-RLHF). For a fair comparison, we use greedy decoding for BLIP+MOCHa as well. As LLaVa-RLHF tends to generate long paragraphs which follow an image description with subjective commentary, we terminate generation after a single sentence, which usually corresponds to an image caption. The instruction given to LLaVa-RLHF is ``describe the image briefly". As seen in the table, our method outperforms LLaVa-RLHF by this measure of open-vocabulary hallucinations. This is further seen in Figure \ref{fig:supp_rlhf}, which shows example captioning predictions for these models, illustrating that LLaVa-RLHF may be more prone to hallucinations.

\input{tables/llava_rlhf_comp}

\input{tables/flickr_eval}
\medskip
\noindent
\label{appendix:generalization}
\textbf{Generalization to Other Datasets and Captioning Styles.} We perform a zero-shot generalization test by evaluating a \methodname{}-tuned model on two additional datasets (different from COCO upon which the model was \methodname{}-tuned). In Table \ref{tab:flickr30k_eval} we can see that the model with \methodname{} fine-tuning shows an improvement in metrics (NLI and BERTScore) over the Flickr30K dataset. Furthermore, we see that non-optimized text quality metrics have similar values between both models, suggesting that \methodname{} tuning generally preserves overall text quality. Supporting this quantitative evaluation, we provide detailed qualitative results on the Flickr30K dataset in the attached visualization tool. In addition, we perform the same test over CC3M. Like earlier, we report a reduced amount of hallucinations - 12.9\% improvement in NLI P(Contradict), while maintaining the same amount of detail (1.44\% improvement in BertScore). Finally, to show that we are able to generalize to other captioning styles, we alter the average caption length by tuning the parameter $\alpha$. The effects of tuning $\alpha$ are presented in table \ref{tab:len_ctrl}.

\input{tables/filtering_analysis_table}
\input{tables/len_ctrl}

\input{figures/rlhf_comparison_supp/rlhf_comp_supp}

%% file: arxiv/tables/results_open_vocab_arxiv.tex
\begin{table*}[t]
  \centering
  \setlength{\tabcolsep}{5pt}
    \begin{tabular}{l|cccccccccccc}
      \toprule
      Model& B@4$\uparrow$ & C$\uparrow$ 
      & {CH$_i$}$\downarrow$ & {CH$_s$}$\downarrow$ & OCH $\downarrow$ & $\bar{p}\downarrow$ & BSc $\uparrow$ &  \\

        \midrule
      
        BLIP-B & 24.8 & 87.5 %
        & 2.6 & 2.8 & 17.6 & 0.206 & 0.557     \\

        BLIP-B+M (ours)
        & \textbf{26.0} & \textbf{91.3} 
        &  \textbf{2.2} & \textbf{2.5} & \textbf{16.4} & \textbf{0.176} & \textbf{0.576}  \\
        
        \midrule

        BLIP-L & 41.5 & 138.4%
        &  2.3 & 3.5 & 19.2 & 0.244 & 0.679     \\

        BLIP-L+M (ours)
        & \textbf{41.9} & \textbf{139.6} %
        &  \textbf{2.1} & \textbf{3.1} & \textbf{18.3} & \textbf{0.206} & \textbf{0.682}  \\

        \midrule
        BLIP2  & 43.4 & \textbf{144.3} %
        & 1.7 & 2.6 & 17.0 & 0.207 & \textbf{0.684}  \\

        BLIP2+M (ours)
        & \textbf{44.0} & \textbf{144.3} %
        & \textbf{1.4} & \textbf{2.3} & \textbf{16.6} & \textbf{0.199} & \textbf{0.684} \\

        \midrule
        GIT-B & 38.7 & 128.1 %
        & 4.2 & 2.9 & 24.7 & 0.284  & 0.656     \\

        GIT-B+M (ours)
        & \textbf{39.0} & \textbf{128.4}
        &  \textbf{3.9} & \textbf{2.7} & \textbf{22.9} & \textbf{0.221} & \textbf{0.657}  \\

        \bottomrule

    \end{tabular}
  \vspace{-5pt}
  \caption{\textbf{Quantitative results} for state-of-the-art VLM models on the COCO Caption Karpathy test set. +M refers to \methodname{}. BSc and $\bar{p}$ denote BERTScore and NLI contradiction probability rewards.  B@4, C, CH, OCH, BSc and $\overline{p}$ denote BLEU-4, CIDEr, CHAIR (i for instance, s for sentence), OpenCHAIR, BERTScore, and NLI $p(\text{contr.})$ metrics respectively. All results are generated by using their officially provided checkpoints and hyperparameters. Best results are shown in \textbf{bold}.
  }
  \label{tab:full_sota_table}
\end{table*}

%% file: tables/additional_ablation_supp.tex
\begin{table*}[t]
  \centering
  \setlength{\tabcolsep}{6pt}
    \begin{tabular}{l|cccccccc}
        \toprule
        Model &  
         OCH $\downarrow$ & B@4$\uparrow$ & C$\uparrow$ 
         & {CH$_i$}$\downarrow$ & {CH$_s$}$\downarrow$ &
        $\bar{p}\downarrow$ & BSc $\uparrow$ & \\
        
        \midrule
        
        BLIP-L & 
         0.270 &  41.5 & 138.4 %
         &  2.3 & 3.5 &  0.244 & 0.679\\

        BLIP-L+M &
         
         0.259  & 41.9 & 139.6 
         & 2.1 & 3.1 & 0.206 &
        0.682   \\

        \midrule
        \hspace{8pt}$-r_f$ &
         
          0.267  & 43.0 & 142.3%
          & 2.8 & 4.4 &
        0.249 & 0.691  \\
        
        \hspace{8pt}$-r_a$ & 
         
         0.257 & 41.1 & 132.9 %
         &  1.5 & 2.3 &
        0.174 & 0.66   \\
        
        \hspace{8pt}$-r_{kl}$ & 
         0.241  & 
          27.6 & 98.9 %
         & 1.4 & 1.9 &  0.135  &0.62   \\
         
        \hspace{8pt}$-ppo$ & 
         0.287 &
         39.4 & 127.6 %
        & 2.5 & 3.76 &  0.212 & 0.664   \\

        \bottomrule
    \end{tabular}

  \vspace{-5pt}
  \caption{\textbf{Additional ablation results.} We ablate the effect of the KL penalty reward $r_{kl}$ and the selection of PPO algorithm. As seen above, removing $r_{kl}$ causes the model to over-optimize the fidelity reward ($\bar{p}$), while replacing PPO with the simpler SCST algorithm (described in Section \ref{sec:supp_additional_abl}) leads to instabilities that degrade performance across metrics.
  }
  \label{tab:additional_ablations_table_supp}
\end{table*}

%% file: tables/results_closed_vocab.tex
\begin{table}[t]
\setlength{\tabcolsep}{3.0pt}
  \begin{tabularx}{\columnwidth}{l|ccccc}
      \toprule
      Model & B@4$\uparrow$  %
      & M$\uparrow$ & C$\uparrow$ & {CH$_s$}$\downarrow$ & {CH$_i$}$\downarrow$\\
      \midrule
      \textbf{Dedicated} \\
      UD-L+Occ\XE & 33.9 %
      & 27.0 & 110.7 & 5.9 & 3.8 \\
      UD-L+Occ\SC & 37.7 %
      & 28.7 & 125.2 & 5.8 & 3.7 \\
      CIIC\XE & 37.3 %
      & 28.5 & 119.0 & 5.3 & 3.6 \\
      CIIC\SC & 40.2 %
      & 29.5 & 133.1 & 7.7 & 4.5 \\
      COSNet\XE & 39.1 %
      & 29.7 & 127.4 & 4.7 & 3.2 \\
      COSNet\SC & \underline{42.0} %
      & 30.6 & \underline{141.1} & 6.8 & 4.2 \\
      \midrule
      \textbf{End-to-end} \\
      BLIP & 41.5 %
      & \underline{31.1} & 138.4 & \underline{3.5} & \underline{2.3} \\
      BLIP-2 %
      & \textbf{43.4} & \textbf{31.7} & \textbf{144.3} & \textbf{2.6}& \textbf{1.7} \\
      \bottomrule
  \end{tabularx}
  
  \caption{\textbf{Older dedicated methods for reduced-hallucination captioning vs. end-to-end modern VLMs for image captioning}. Results are given on the Karpathy test split of MS-COCO dataset, including closed-vocabulary hallucination metrics as commonly reported by such dedicated methods. B@4, C, M, CH denote BLEU-4, CIDEr, METEOR, and CHAIR metrics respectively. We see that older, dedicated methods with weaker backbones are outperformed by modern VLMs on all metrics, including the smaller BLIP(-Large) and the larger BLIP-2(-2.7B). XE and SC indicate cross-entropy and SCST (RL) optimization respectively. Best and second-best metric values are shown in \textbf{bold} and \underline{underlined text} respectively.}
  \label{tab:closed_vocab}
  \vspace{-0.19in}
\end{table}

%% file: figures/openchair_coverage_supp/och_coverage.tex
\begin{figure}
    \centering
    
    \includegraphics[trim={0.3cm 0.2cm 0cm 0cm},clip,width=0.45\textwidth]{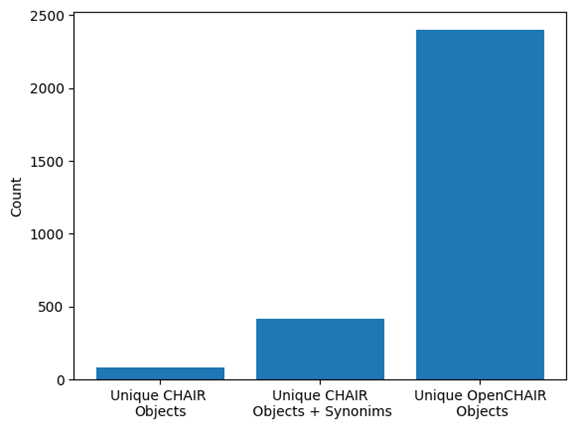}
    \caption{\textbf{Object \textit{Type} Coverage, CHAIR vs. \metricname{}.} We display the object type coverage of CHAIR (over MS-COCO) and \metricname{}, measured as the number of unique objects. In OPENChair, objects are found using the parsing method described in Section \ref{sec:och_details_supp}. As can be observed, the proposed benchmark has significantly greater coverage of different objects.
    } 
    \label{fig:och_cov}
\end{figure}

%% file: tables/efficient_och.tex
\begin{table}[t]
  \centering
  \setlength{\tabcolsep}{2pt}
    \begin{tabular}{l|ccccccccc}
        \toprule
        LLM Used & Blip-B & Blip-L
        & Blip-2 & Git-B  \\
        
        \midrule
        
        LLaMA-2-70B & 16.53 & 8.05 & 4.29 & 9.59 \\
        LLaMA-3-8B & 16.64 & 6.19 & 2.74 & 9.85 \\
        
        \bottomrule
    \end{tabular}

  \caption{\textbf{\metricname{} Efficiency. } We show that \metricname{} can be computed with smaller, more efficient LLMs like LLaMA-3-8B-Instruct. In each entry we evaluate the relative improvement in \% in \metricname{} scores for a \methodname{}-tuned model, as computed by the respective LLM. We observe that both the small and large LLMs correlate well on their evaluation, and capture the same trends of model improvement.}
  \label{tab:efficient_och}
\end{table}

%% file: tables/results_abl.tex
\begin{table}[t]
  \centering
  \setlength{\tabcolsep}{1.8pt}
    \begin{tabular}{l|cccccccccccc}
        \toprule
        Model& B@4$\uparrow$ & C$\uparrow$ 
        & {CH$_i$}$\downarrow$ & {CH$_s$}$\downarrow$ &
        $\bar{p}\downarrow$ & BSc $\uparrow$ &  \\
        
        \midrule
        
        BLIP & 41.5 & 138.4 %
        & 2.3 & 3.5 &
        0.246 & 0.679 \\
        BLIP+M & 41.9 & 139.6 %
        & 2.1 & 3.1 &
        0.206 & 0.682  \\

        \midrule
        \hspace{8pt}$-r_f$ & 43.0 & 142.3 
        & 2.8 & 4.4 &
        0.249 & 0.691     \\
        \hspace{8pt}$-r_a$ & 41.1 & 132.9 
        &  1.5 & 2.3 &
        0.174 & 0.66     \\

        \bottomrule
    \end{tabular}

  \vspace{-0.1in}
  \caption{\textbf{Reward Ablation.} We ablate the effect of the fidelity $r_f$ and adequacy $r_a$ terms in our reward function, finding that using each alone significantly degrades performance with respect to hallucinations or textual quality.
  }
  
  \label{tab:abl_table}
\end{table}

%% file: figures/qualitative/reward_ablation_figure_multiimage.tex
\begin{figure}[ht]
    \begin{minipage}{\linewidth}
        \begin{tabularx}{\linewidth}[t]{|s|R|R|}
            \hline
            \multicolumn{1}{|c|}{\raisebox{0.077\textwidth}{\includegraphics[width=0.07\textwidth]{figures/qualitative/reward.png}}}
            & \lapbox[\width]{-0.57em}{\raisebox{-1.12mm}{\includegraphics[trim={0 1pt 0pt 6pt},clip,height=1.8cm]{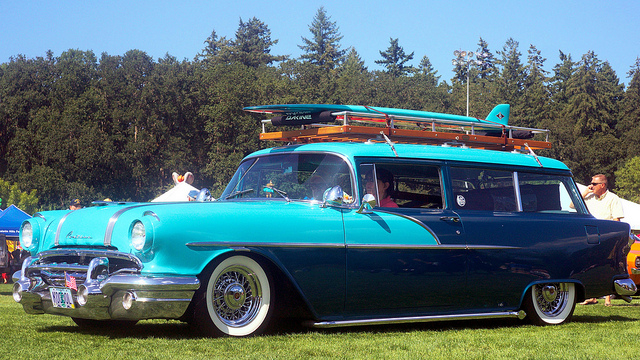}}}
            & \lapbox[\width]{-0.65em}{\raisebox{-1.14mm}{\includegraphics[trim={4pt 57pt 0 55pt},clip,height=1.8cm]{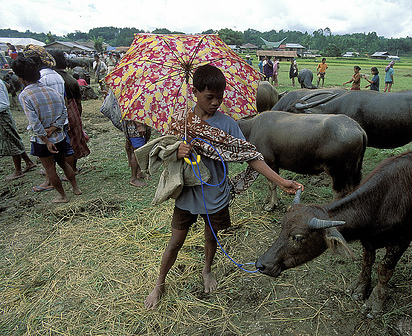}}}
            \\
            \hline
            \multicolumn{1}{|c|}{\raisebox{-0.07\textwidth}{$\emptyset$}}
            & \small{\emph{This is a picture of a large old fashioned car that was parked by a group of people}}
            & \small{\emph{People at festival standing around in open field}} \\ \hline
            
            \multicolumn{1}{|c|}{\raisebox{-0.05\textwidth}{$-r_f$}}
            & \small{\emph{A car parked in the grass with a surfer standing near it}}
            & \small{\emph{A woman standing next to a herd of animals with an umbrella}} \\ \hline

            \multicolumn{1}{|c|}{\raisebox{-0.05\textwidth}{$-r_a$}}
            & \small{\emph{Spectators could enjoy the old fashions of the fifties}}
            & \small{\emph{That are some very nice people who are very fun to view them}} \\ \hline
            
            \multicolumn{1}{|c|}{\raisebox{-0.05\textwidth}{$r$}}
            & \small{\emph{A vintage car parked on a field next to people}}
            & \small{\emph{A young man with a large umbrella next to a herd of animals}} \\ \hline
            
        \end{tabularx}
    \end{minipage}
    \vspace{-4.3pt}
    \caption{\textbf{Ablating our multi-objective reward function}. Above we show captions sampled from models with different reward functions. Top row depicts the initial model (before optimization).
    As can be seen in the table, generations of the base model ($\emptyset$) and the model trained without the fidelity objective ($-r_f$) contain various hallucinations that contradict the image, like stating that the car was \emph{parked by a group of people}, confusing between an ordinary person and a \emph{surfer}, and stating that the boy is a \emph{woman}. In contrast, those from the model without the adequacy objective ($-r_a$) are generic and neutral with respect to the image (without explicitly contradicting it), e.g. the abstract statement about the \emph{spectators enjoying the old fashions of the fifties}. At last, optimizing for both ($r$) yields captions that are both descriptive and consistent with the input condition, similar to the reference captions\setcounter{footnote}{1}\protect\footnotemark{} that were provided by human annotators.}
    \label{fig:qual_reward_abl}
\end{figure}

\footnotetext{Reference ground truth captions: \emph{A car with some surfboards in a field} (left) and \emph{A boy holding umbrella while standing next to livestock} (right).}

%% file: tables/llava_rlhf_comp.tex
\begin{table}[t]
  \centering
  \setlength{\tabcolsep}{5pt}
    \begin{tabular}{l|c}
        \toprule
        Model & OCH $\downarrow$  \\
        
        \midrule 
        LLaVa-RLHF\textsubscript{S} & 0.396  \\
        LLaVa-RLHF\textsubscript{G} & 0.401 \\
        BLIP-L+M\textsubscript{G} & \textbf{0.360}\\

        \bottomrule
    \end{tabular}

  \caption{OPENChair comparison between LLaVa-RLHF and BLIP-L+\methodname{} over 100 random samples. For LLaVa-RLHF, S stands for stochastic sampling with default parameters, and G stands for greedy decoding (as beam search is not implemented for LLaVa-RLHF). For fair comparison, we also apply greedy decoding to BLIP-L+\methodname{}.
  }
  \label{tab:llava_rlhf_comp}
\end{table}

%% file: tables/flickr_eval.tex
\begin{table}[t]
  \centering
  \setlength{\tabcolsep}{2pt}
    \begin{tabular}{l|cccccccccc}
        \toprule
        Model& B@4$\uparrow$ & C$\uparrow$ 
        & $\bar{p}\downarrow$ & BSc $\uparrow$ &  \\
        
        \midrule
        
        BLIP & \textbf{29.0} & 73.2 
        & 0.335 & 0.603 \\
        BLIP+M & 28.9 & \textbf{73.6} %
        & \textbf{0.296} & \textbf{0.607}  \\
        
        \bottomrule
    \end{tabular}

  \caption{\textbf{Evaluation over Flickr30K dataset.} We perform a zero-shot evaluation of BLIP-Large with and without \methodname{} (performed on COCO) on an additional dataset. As seen above, improvements to the optimized metrics ($\bar{p}$ and BERTScore) transfer to the new dataset, while other text quality metrics have similar values before and after \methodname{}-tuning, suggesting that overall text quality is generally preserved.
  }
  \label{tab:flickr30k_eval}
\end{table}

%% file: tables/filtering_analysis_table.tex
\begin{table}[ht]
  \centering
  \setlength{\tabcolsep}{3pt}
    \begin{tabular}{l|c|c}
        \toprule
              & \multicolumn{2}{c}{\methodname{}'s Improvement (OCH) in \%}  \\
           Model   &     \hspace{0.25cm} without filtering \hspace{0.25cm} & with filtering   \\
        
        \midrule
        
        BLIP-B    & 4.9\%   &  4.8\%   \\
        BLIP-L    & 2.0\%   &  2.3\%   \\
        BLIP2     & 7.3\%   &  6.9\%   \\
        GIT-B     & 7.0\%   &  7.1\%   \\

        \bottomrule
        
    \end{tabular}

\vspace{-5pt}
  \caption{\textbf{Performance of \methodname{} with and without manual filtering}.
  We compare performance on the \metricname{} (OCH) benchmark before and after it is manually filtered, as measured by the improvement provided by \methodname{} on \metricname{} scores across various models. We observe similar results before and after filtering, corresponding to the relative high quality of the generated data and consistent with the small proportion of data that was removed.
  }
  \label{tab:och_filtering_analysis}
\end{table}

%% file: tables/len_ctrl.tex
\begin{table}[ht]
  \centering
  \setlength{\tabcolsep}{2pt}
    \begin{tabular}{c|ccccc}
        \toprule
        $\boldsymbol\alpha$ & 0 & 0.25 & 0.5 & 0.75 & 1 \\
        
        \midrule
        
        \textbf{Mean Word Count} & 20.4 & 15.1 & 12.8 & 11.3 & 9.7  \\
        
        \bottomrule
    \end{tabular}

  \caption{\textbf{Controlling caption length.} By tuning the parameter $\alpha$ we are able to train captioners with different output lengths. Results are show for Blip-Large over MS-COCO.
  }
  \label{tab:len_ctrl}
\end{table}

%% file: figures/rlhf_comparison_supp/rlhf_comp_supp.tex
\begin{figure}[ht!]
    \newcolumntype{q}{>{\hsize=.185\hsize\RaggedRight}X}
    \newcolumntype{Q}{>{\hsize=0.005\hsize\RaggedRight}X}
    \setlength{\tabcolsep}{0.02pt}
    \setlength{\extrarowheight}{5pt} %
    \begin{tabularx}{\textwidth}{cp{0.3cm}qq}
        && \centering \textbf{LLaVa-RLHF} & \multicolumn{1}{c}{\textbf{BLIP-L+\methodname{}}} \\
        \raisebox{-0.84\height}{\includegraphics[width=0.2\linewidth]{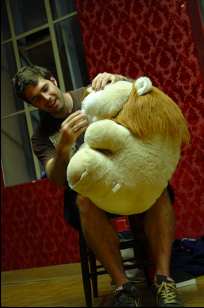}} && 
        {\small \emph{A man sitting on a chair with a stuffed animal, \hlcr{specifically a teady bear}, on his lap} } &
        {\small \emph{a man sitting on a chair holding a large stuffed animal}}
    \end{tabularx}

    \vspace{-0.05in}
    \caption{\textbf{LLaVa-RLHF vs. \methodname{}}. We illustrate that RLHF training does not necessarily solve the hallucination problem of VLM models by showing a generation produced by LLaVa-RLHF~\cite{sun2023llavarlhf} compared to BLIP+\methodname{}. For both models, we use the prompt \emph{``a photography of"} for generation. See Table \ref{tab:llava_rlhf_comp} for a quantitative comparison.}
    \vspace{-0.15in}
    \label{fig:supp_rlhf}
\end{figure}

%% file: supp_03_extended_rw.tex
\section{Extended Discussion of Previous Work}

We provide here an extended discussion of related methods, shown in Figure \ref{fig:taxonomy}.

\subsection{Similarity Based Metrics}
CLIPScore~\cite{hessel2022clipscore} propose CLIP cross-modal similarity for detecting mismatches between text and images, including hallucinations, and \citet{shi2022emscore} propose a similar embedding-based metric for video captioning. However, \citet{xu2023challenges} find that CLIP tends to assign high similarity to texts with minor modifications (``hard negatives'') that contradict the corresponding image. The Egoshots Semantic Fidelity metric~\cite{agarwal2020egoshots} and VIFIDEL~\cite{madhyastha-etal-2019-vifidel} use embedding similarity between object annotations or detections in images and items in predicted captions. FAIEr~\cite{wang2021faier} proposes a learned fidelity metric, which must be trained on automatically-generated scene graphs. Unlike these methods, our benchmark provides an explicit measure of hallucinations that can be directly examined (predicted captions on the \metricname{} benchmark images).

\subsection{Closed Vocabulary Algorithms}

UD-L~\cite{biten2021clockonthebeach} identifies object hallucinations with bias towards the prior distribution of objects in context found in the training data, and proposes the use of synthetically debiased captions. 
CIIC~\cite{liu2022ciic} focuses on captioning models with a closed-vocabulary object detection backbone, inserting components into the object detector and text decoder to reduce spurious correlations. TLC~\cite{petryk2023tlc} proposes a text decoding method applied to existing captioning models, to avoid generating COCO object tokens if they have insufficient confidence.
The more recent work ObjMLM~\cite{Dai2023Plausible} proposes masking objects from closed vocabulary lists as a training objective.
The concurrent work Woodpecker~\cite{yin2023woodpecker} combines closed-vocabulary object detection with LLM-guided decoding to avoid hallucinations in generated text.
Unlike these works, our \methodname{} optimization method does not rely on a closed list of object types.